# Variance Reduced Policy Gradient Method for Multi-Objective Reinforcement Learning

**Davide Guidobene**[*1], **Lorenzo Benedetti**[*1],
**Diego Arapovic**[*1]

[1]Department of Computer Science
ETH Zurich

**Abstract**

Multi-Objective Reinforcement Learning (MORL) is a generalization of traditional Reinforcement Learning (RL) that aims to optimize multiple, often conflicting objectives simultaneously rather than focusing on a single reward. This approach is crucial in complex decision-making scenarios where agents must balance trade-offs between various goals, such as maximizing performance while minimizing costs. We consider the problem of MORL where the objectives are combined using a non-linear scalarization function. Just like in standard RL, policy gradient methods (PGMs) are amongst the most effective for handling large and continuous state-action spaces in MORL. However, existing PGMs for MORL suffer from high sample inefficiency, requiring large amounts of data to be effective. Previous attempts to solve this problem rely on overly strict assumptions, losing PGMs' benefits in scalability to large state-action spaces. In this work, we address the issue of sample efficiency by implementing variance-reduction techniques to reduce the sample complexity of policy gradients while maintaining general assumptions.

## Introduction

Traditional RL focuses on agents learning to make decisions by interacting with an environment to maximize a cumulative reward. However, many real-world problems, ranging from Healthcare and Treatment Optimization (Huo and Tang 2022) to Reinforcement Learning from Human Feedback (Rame et al. 2024), are inherently multi-objective, where optimizing for one objective may lead to suboptimal outcomes in others. MORL addresses this complexity by providing mechanisms for learning policies that consider multiple criteria, potentially finding Pareto-optimal solutions that best compromise among them.

The setting of MORL is intrinsically more complex than traditional reinforcement learning due to the challenge of balancing multiple objectives with a non-linear scalarization function. Policy-gradient methods (PGMs) are among the most effective approaches for MORL in large state-action spaces. However, a significant downside is their high sample complexity due to high variance in the gradient estimates. Recently, Bai, Agarwal, and Aggarwal (2022) proposed a Policy Gradient Algorithm with a sample complexity of $\tilde{O}(\frac{M^4}{\varepsilon^4})$,

---

[*]These authors contributed equally.

where $M$ represents the number of objectives and $\varepsilon$ the error tolerance.

Later, Zhou et al. (2022) tried to improve on this result proposing a fast PGM, with a convergence rate as low as $\tilde{O}(\frac{1}{T})$. However, guarantees of their algorithm rely on a parameterization that scales linearly with the size of the state-action space. Therefore, it is not well suited for large state-action spaces.

We aim to tackle the issue of sample complexity of PGMs in MORL while keeping their important advantage of scaling in large state-action spaces. We accomplish this goal by applying variance reduction to policy gradients, similarly to Zhang et al. (2021), and we propose a novel algorithm improving over current state-of-the-art (Bai, Agarwal, and Aggarwal 2022).

**Main results.** The main contributions of this work are:

- We propose the MO-TSIVR-PG algorithm. Compared to the TSVIR-PG algorithm presented in Zhang et al. (2021), we are able to implement variance education without the need to explicitly keep track of the occupancy measure $\lambda$ by leveraging the structure of MORL settings. This fundamental change in the structure of the algorithm, allows it to operate on large state-action spaces.

- Furthermore, we relax the assumption of softmax parameterization of the policy used in Zhang et al. (2021), allowing for a more general class of parameterizations, including the Gaussian policy. Therefore, theoretical results are more general and are not bound to the softmax parameterization, which would not allow the algorithm to operate in continuous action spaces.

- We relax the assumption made by Bai, Agarwal, and Aggarwal (2022) that the variance of the gradient estimator is bounded by a constant factor. Without this assumption, the variance may scale linearly with the number of objectives. Given that the theoretical analysis in this paper aims to provide bounds with respect to the optimality gap $\varepsilon$ and the number of objectives $M$, we consider this assumption to be overly restrictive.

- We analyze the theoretical guarantees of MO-TSIVR-PG, showing that it finds a $\varepsilon$-stationary policy with $\tilde{\mathcal{O}}(\frac{M^3}{\varepsilon^3})$ samples. Moreover, if the scalarization function is concave and the policy parameterization is overparameterized, we show that the algorithm achieves $\varepsilon$-function optimality gap with $\tilde{\mathcal{O}}(\frac{M^5}{\varepsilon^2})$ samples.
- To provide a stable comparison, we analyze Bai et a.'s algorithm under the same assumption as MO-TSIVR-PG. Crucially, our analysis does not require Assumption 5 from Bai, Agarwal, and Aggarwal (2022). We obtain sample complexities of $\tilde{\mathcal{O}}(\frac{M^4}{\varepsilon^4})$ and $\tilde{\mathcal{O}}(\frac{M^6}{\varepsilon^3})$ for stationary convergence and global convergence with concave scalarization, respectively. In both cases, our algorithm has better theoretical guarantees.
- Finally, we show that MO-TSIVR-PG outperforms Bai et al's algorithm experimentally under every tested setting.

## Problem Formulation

**Problem Setting**
We consider an infinite horizon MDP defined as the tuple $(\mathcal{S}, \mathcal{A}, \mathbb{P}, r_{1:M}, \gamma, \rho)$, where $\mathcal{S}$ and $\mathcal{A}$ represent the finite state and action sets respectively, $\mathbb{P} : \mathcal{S} \times \mathcal{A} \to \Delta(\mathcal{S})$ is the transition probability, $\rho \in \Delta(\mathcal{S})$ is the initial state distribution, and $\gamma \in (0, 1)$ is the discount factor. In our scenario, the departure from the standard RL framework lies in the nature of the rewards, which are not single scalar values but rather $M$-dimensional vectors. Formally, this is expressed as $r_m : \mathcal{S} \times \mathcal{A} \to \mathbb{R}^M$, where $m$ ranges over the indices $[M]$. The agent adopts a policy $\pi : \mathcal{S} \to \Delta^{|\mathcal{A}|}$ to choose what action to take. Then, we consider the standard and truncated value functions for all $M$ rewards, defined by

$$J_m(\pi) := \mathbb{E}_{\rho(s_0), \pi(a_t|s_t), \mathbb{P}(s_{t+1}|s_t, a_t)} \left[ \sum_{t=0}^{\infty} \gamma^t r_m(s_t, a_t) \right]$$

$$J_m^H(\pi) := \mathbb{E}_{\rho(s_0), \pi(a_t|s_t), \mathbb{P}(s_{t+1}|s_t, a_t)} \left[ \sum_{t=0}^{H-1} \gamma^t r_m(s_t, a_t) \right]$$

and we define $\Omega$ to be the range of $\mathbf{J}$.

**Problem Goal**
We consider a policy $\pi_\theta$ parameterized by some $\theta$. The agent aims to maximize the scalarization $f : \Omega \to \mathbb{R}$, also called the joint objective function. Formally

$$\max_{\pi_\theta} f(\mathbf{J}(\pi_\theta))$$

where $\mathbf{J}(\pi_\theta) := \mathbf{J}(\theta) := (J_1(\pi_\theta), \ldots, J_M(\pi_\theta))^\top$.

**Assumptions**
**Assumption 1**
The absolute value of $r_m$ is bounded $\forall m \in [M]$. Without loss of generality, we assume $r_m \in [0, 1] \forall m \in [M]$.

As a trivial consequence of Assumption 1, $\Omega = [0, \frac{1}{1-\gamma}]^M$.

**Assumption 2**
The log-likelihood function of the policy parameterization is twice differentiable, $G$-Lipschitz and $B$-smooth. Formally

$$\|\nabla_\theta \log \pi_\theta(a|s)\| \leq G$$
$$\|\nabla_\theta^2 \log \pi_\theta(a|s)\| \leq S$$

for all state-action pairs $(s, a) \in \mathcal{S} \times \mathcal{A}$.

Notice that both the softmax and the Gaussian policy, satisfy those properties. Therefore, compared to the corresponding assumption on the policy parameterization in Zhang et al. (2021), this allows for more general results. Observe that relaxing this assumption is of particular importance because a restriction to the softmax parameterization would not allow our algorithm to scale to large and continuous action spaces.

The following Assumption corresponds exactly to Assumption 4 from Bai, Agarwal, and Aggarwal (2022).

**Assumption 3**
$f \in C^1$ and all partial derivatives of $f$ are assumed to be locally $L_f$-Lipschitz functions, formally

$$\|\nabla_\mathbf{J} f(\mathbf{J}') - \nabla_\mathbf{J} f(\mathbf{J}^\dagger)\|_\infty \leq L_f \|\mathbf{J}' - \mathbf{J}^\dagger\| \quad \forall \mathbf{J}', \mathbf{J}^\dagger \in \Omega.$$

**Lemma 1**
The partial derivatives are bounded by a constant $C > 0$ on $\Omega$, formally

$$\|\nabla_\mathbf{J} f(\mathbf{J}')\|_\infty \leq C \quad \forall \mathbf{J}' \in \Omega.$$

(Lemma 1 of Bai, Agarwal, and Aggarwal (2022)).

**Lemma 2**
Let assumptions 1, 2 and 3 be satisfied, the following results hold:
(i) The function $f(\mathbf{J}(\cdot))$ is smooth wrt $\theta$ with parameter $L_\theta$

$$L_\theta := \frac{MCS}{(1-\gamma)^2}.$$

(ii)
$$\|\nabla_\theta f(\mathbf{J}(\theta)) - \nabla_\theta f(\mathbf{J}^H(\theta))\| \leq D_\mathbf{J} \gamma^H$$

where
$$D_\mathbf{J} := \frac{MG}{(1-\gamma)^2} \left[ \sqrt{M} L_f \frac{1 - \gamma^H - H\gamma^H(1-\gamma)}{1-\gamma} + \right.$$
$$\left. + C[1 + H(1-\gamma)] \right].$$

(Lemmas 6 and 7 of Bai, Agarwal, and Aggarwal (2022)).

The following two Assumptions, replace assumptions appearing in similar works using variance reduction in the literature such as Assumptions 5.10 and 5.11 of Zhang et al. (2021) and Assumptions 4.9 and 4.10 of Barakat, Fatkhullin, and He (2023). They are needed for the global convergence Theorems presented later. However, notice that our assumptions are strictly more general than the ones in the above mentioned papers.

**Assumption 4**
$f$ is a concave function of the expected sum of discounted rewards.

**Assumption 5**
The policy parameterization $\theta \to \pi_\theta$ overparameterizes the set of policies, more specifically we have:
1. For any $\theta \in \Theta$, there exist neighborhoods $\theta \in \mathcal{U}_\theta \subset B(\theta, \delta)$ and $\mathbf{J}(\theta) \in \mathcal{V}_{\mathbf{J}(\theta)} \subset \mathbf{J}(B(\theta, \delta))$ such that the restriction $\mathbf{J}|_{\mathcal{U}_\theta}$ is a bijection from $\mathcal{U}_\theta$ to $\mathcal{V}_{\mathbf{J}(\theta)}$. Moreover, the inverse $(\mathbf{J}|_{\mathcal{U}_\theta})^{-1}$ is $l_\theta$-Lipschitz continuous for any $\theta \in \Theta$.
2. Let $\pi_{\theta^*}$ be the optimal policy. Moreover, there exists $\bar{\varepsilon}$ such that $(1-\varepsilon)\mathbf{J}(\theta) + \varepsilon \mathbf{J}(\theta^*)$ for all $\varepsilon \leq \bar{\varepsilon}$ and for any $\theta$.

# Multi-Objective Truncated Stochastic Incremental Variance-Reduced Policy Gradient

In the following, $\tau$ represents a trajectory sampled from the MDP following policy $\pi_{\theta_1}$.

## Importance Sampling

The technique used to achieve variance reduction makes use of IS weights defined by

$$\omega_t(\tau|\theta_1, \theta_2) = \prod_{h=0}^{t} \frac{\pi_{\theta_2}(a_h|s_h)}{\pi_{\theta_1}(a_h|s_h)}.$$

In order to bound the variance of our estimators, we need to first establish a bound on the variance of the IS weights.

## Lemma 3

Let assumption 2 be satisfied and assuming $\|\theta_1 - \theta_2\| \leq \delta$, then

$$\mathbb{E}[\omega_t(\tau|\theta_1, \theta_2)] = 1$$

$$\mathbb{V}[\omega_t(\tau|\theta_1, \theta_2)] \leq C_\omega(t)\|\theta_1 - \theta_2\|^2$$

where $C_\omega(t) := t(2tG^2 + S)(\exp\{2GH\delta\} + 1), \forall t \in \{0, \ldots, H-1\}$.

Notice that this Lemma is our version of Lemma 5.7 from Zhang et al. (2021). The differences in the constants are due to our different assumptions in the policy parameterization.

## Estimators

We consider the following off-policy version of the estimator proposed by (Bai, Agarwal, and Aggarwal 2022) for the Policy Gradient $\nabla_\theta f(\mathbf{J}(\theta_2))$:

$$g(\tau|\theta_1, \theta_2, \hat{\mathbf{J}}) = \sum_{t=0}^{H-1} \omega_t(\tau|\theta_1, \theta_2) \nabla_\theta \log \pi_{\theta_2}(a_t^i|s_t^i) \cdot$$

$$\cdot \left( \sum_{m=1}^{M} \frac{\partial f}{\partial J_m}(\hat{\mathbf{J}}) \left( \sum_{h=t}^{H-1} \gamma^h r_m(s_h^i, a_h^i) \right) \right).$$

Moreover, we also define an estimator for the sum of discounted rewards as follows:

$$\mathbf{J}(\tau|\theta_1, \theta_2) = \sum_{t=0}^{H-1} \gamma^t \omega_t(\tau|\theta_1, \theta_2) \mathbf{r}(s_t, a_t)$$

where $\mathbf{r}(s, a) = (r_1(s, a), \ldots, r_M(s, a))$.

Notice that $\omega_t(\tau|\theta, \theta) \equiv 1$. Thus, for ease of notation, we will shorten $g(\tau|\theta, \theta, \hat{\mathbf{J}}) = g(\tau|\theta, \hat{\mathbf{J}})$ and $\mathbf{J}(\tau|\theta) = \mathbf{J}(\tau|\theta_1, \theta_2)$ in the algorithms.

## Algorithms

Algorithm 1 is the algorithm presented in Bai, Agarwal, and Aggarwal (2022) choosing $N = |\mathcal{N}_1^i| = |\mathcal{N}_2^i|$. To improve on its sample complexity, we present Algorithm 2, implementing variance reduction techniques analogous to the ones introduced by Zhang et al. (2021).

It is important to notice that our algorithm presents three key differences from TSIVR-PG (Zhang et al. 2021), which make it particularly suitable for the setting of Multi-Objective Reinforcement Learning:

---

**Algorithm 1: MO-PG**

**Input:** initial point $\theta_0$, batch size $N$, trajectory length $H$, stepsize $\eta$, number of epochs $T$.
**for** Epoch $i = 0, \ldots, T-1$ **do**
    Sample $N$ trajectories of length $H$ following policy $\pi_{\theta_i}$ and collect them as $\mathcal{N}_1^i$.
    Estimate $J_i = \frac{1}{N} \sum_{\tau \in \mathcal{N}_1^i} \mathbf{J}(\tau|\theta_i)$.
    Sample $N$ trajectories of length $H$ following policy $\pi_{\theta_i}$ and collect them as $\mathcal{N}_2^i$.
    Estimate $g_i = \frac{1}{N} \sum_{\tau \in \mathcal{N}_2^i} g(\tau|\theta_i, J_i)$.
    $\theta_{i+1} = \theta_i + \eta g_i$
**end for**

---

**Algorithm 2: MO-TSIVR-PG**

**Input:** initial point $\theta_0^1 = \tilde{\theta}_0$, batch sizes $B$ and $N$, trajectory length $H$, stepsize $\eta$, epoch length $m$, epochs $T$, gradient truncation radius $\delta$.
**for** Epoch $i = 0, \ldots, T-1$ **do**
    **for** Iteration $j = 0, \ldots, m-1$ **do**
        **if** $j = 0$ **then**
            Sample $N$ trajectories of length $H$ following policy $\pi_{\theta_0^i}$ and collect them as $\mathcal{N}_1^i$.
            Estimate $J_0^i = \frac{1}{N} \sum_{\tau \in \mathcal{N}_1^i} \mathbf{J}(\tau|\theta_0^i)$.
            $P_0^i = \mathbf{Proj}_\Omega(J_0^i)$
            Sample $N$ trajectories of length $H$ following policy $\pi_{\theta_0^i}$ and collect them as $\mathcal{N}_2^i$.
            Estimate $g_0^i = \frac{1}{N} \sum_{\tau \in \mathcal{N}_2^i} g(\tau|\theta_0^i, P_0^i)$.
        **else**
            Sample $B$ trajectories of length $H$ following policy $\pi_{\theta_j^i}$ and collect them as $\mathcal{B}_{j,1}^i$.
            $J_j^i = \sum_{\tau \in \mathcal{B}_{j,1}^i} \frac{\mathbf{J}(\tau|\theta_j^i) - \mathbf{J}(\tau|\theta_j^i, \theta_{j-1}^i)}{B} + J_{j-1}^i$
            $P_j^i = \mathbf{Proj}_\Omega(J_j^i)$
            Sample $B$ trajectories of length $H$ following policy $\pi_{\theta_j^i}$ and collect them as $\mathcal{B}_{j,2}^i$.
            $g_j^i = \sum_{\tau \in \mathcal{B}_{j,2}^i} \frac{g(\tau|\theta_j^i, P_j^i) - g(\tau|\theta_j^i, \theta_{j-1}^i, P_{j-1}^i)}{B} + g_{j-1}^i$
        **end if**
        $\theta_{j+1}^i = \mathbf{Proj}_{B(\theta_j^i, \delta)}\left(\theta_j^i + \eta g_j^i\right)$
    **end for**
    Set $\theta_0^{i+1} = \tilde{\theta}_i = \theta_m^i$.
**end for**

1. The gradient estimator uses an estimate of $\mathbf{J}(\theta_j^i)$ directly instead of estimating the state occupancy measure $\lambda(\theta_j^i)$. This allows our algorithm to work on large and continuous state-action spaces.

2. At every iteration, we sample two batches of trajectories instead of just one. This allows us to estimate the gradient and the sum of discounted rewards independently, leading to a better bound in Lemma 5 and, consequently, to a better dependency on the number of objectives in Theorem 2.

3. The estimate of $\mathbf{J}$ used by the gradient estimator $g(\tau|\theta_1,\theta_2,\cdot)$ is $P_j^i$, i.e. the projection of $J_j^i$ on $\Omega$. This allows for more natural assumptions on $f$, which does not need to be defined outside the domain $\Omega$. At the same time, the use of $P_j^i$ prevents importance sampling from giving an unrealistic estimate of $\mathbf{J}$, outside its range.

## Analysis and Sample Efficiency of MO-PG and MO-TSVIR-PG

In the following, we analyze and compare the sample complexity of Algorithms 1 and 2 under the same assumptions.

Bai, Agarwal, and Aggarwal (2022) assume the variance of $g(\tau|\theta_1,\theta_2,\mathbf{J}(\theta_2))$ is bounded by a constant factor. However, under our assumptions, we can only claim a bound that depends on the number of objectives. For this reason, we think Assumption 5 from their paper is too strong and our analysis is fundamentally different, following a scheme closer to the one used by Zhang et al. (2021). Consequently, we provide a new analysis for Algorithm 1, instead of simply citing the results from Bai, Agarwal, and Aggarwal (2022) allowing a more fair comparison with our algorithm.

Notice that we choose $N = |\mathcal{N}_1^i| = |\mathcal{N}_2^i|$. This choice simplifies the analysis. Furthermore, according to our calculation, choosing different batch sizes would not result in a better sample complexity.

### Bounding the Variance

We present two results bounding the variance of the gradient estimates for Algorithms 1 and 2, respectively:

**Lemma 4**
Let Assumptions 1, 2 and 3 hold. For the gradient estimates $g_i$ of Algorithm 1, it holds

$$\mathbb{E}\left[\|g_i - \boldsymbol{\nabla}_\theta f(\mathbf{J}(\theta_i))\|^2\right] \leq \frac{C_1}{N} + C_2\gamma^{2H}$$

for some $C_1, C_2 \in \mathcal{O}(M^3)$, whose explicit expression is derived in the Appendix.

**Lemma 5**
Under the same assumptions of Lemma 4, for the gradient estimates $g_j^i$ of Algorithm 2, it holds

$$\mathbb{E}\left[\|g_j^i - \boldsymbol{\nabla}_\theta f(\mathbf{J}(\theta_j^i))\|^2\right] \leq \frac{C_1}{N} + C_2\gamma^{2H} + \\ + \frac{C_3}{B}\sum_{j'=1}^{j}\mathbb{E}\left[\|\theta_{j'-1}^i - \theta_{j'}^i\|^2\right]$$

for some $C_1, C_2, C_3 \in \mathcal{O}(M^3)$, whose explicit expression is derived in the Appendix. $C_1$ and $C_2$ are identical to those in Lemma 4.

Notice that Lemma 5, compared to Lemma 4, has an extra term that comes from the variance of the importance sampling weights in our estimators. On the other hand, compared to Lemma 5.8 of Zhang et al. (2021), we are able to get rid of one term in the bound of Lemma 5. This improvement is due to the fact that our algorithm samples additional trajectories to compute the estimate $J_j^i$. This results in a better dependency on the number of objectives in the sample complexity in Theorem 2.

### Stationary Convergence

We present two theorems that establish the sample complexity required for stationary convergence. Notice that the analysis we conduct to obtain these theoretical guarantees still only requires Assumptions 1, 2, and 3.

Using the result from Lemma 4, our analysis gives the following result for MO-PG.

**Theorem 1**
Let Assumptions 1, 2 and 3 hold. Choosing $T = M\varepsilon^{-2}, N = M^3\varepsilon^{-2}, H = \ln(M\varepsilon^{-1})$ and $\eta = \frac{1}{2L_\theta}$, Algorithm 1 achieves $\mathbb{E}\left[\|\boldsymbol{\nabla}_\theta f(\mathbf{J}(\theta_{out}))\|\right] \leq \tilde{\mathcal{O}}(\varepsilon)$ with sample complexity

$$T \times (2N) \times H = \tilde{\mathcal{O}}(\frac{M^4}{\varepsilon^4})$$

where $\theta_{out}$ is sampled uniformly at random from $\{\theta_i\}_{i=0,\ldots,T-1}$.

Analogously, starting from Lemma 5, we obtain the following result for MO-TSIVR-PG.

**Theorem 2**
Under the same assumptions of Theorem 1, choosing $T = \varepsilon^{-1}, m = B = M^{3/2}\varepsilon^{-1}, N = B^2 = M^3\varepsilon^{-2}, H = \ln(M\varepsilon^{-1})$ and $\eta = \frac{1}{1+C_3/(ML_\theta^2)}\cdot\frac{1}{2L_\theta}$, Algorithm 2 achieves $\mathbb{E}\left[\|\boldsymbol{\nabla}_\theta f(\mathbf{J}(\tilde{\theta}_{out}))\|\right] \leq \tilde{\mathcal{O}}(\varepsilon)$ with sample complexity

$$T \times (2N + 2(m-1)B) \times H = \tilde{\mathcal{O}}(\frac{M^3}{\varepsilon^3})$$

where $\tilde{\theta}_{out}$ is sampled uniformly at random from $\{\theta_j^i\}_{j=0,\ldots,m-1}^{i=1,\ldots,T}$.

Comparing the two results, our analysis shows that our algorithm offers better theoretical guarantees than the one proposed by Bai, Agarwal, and Aggarwal (2022).

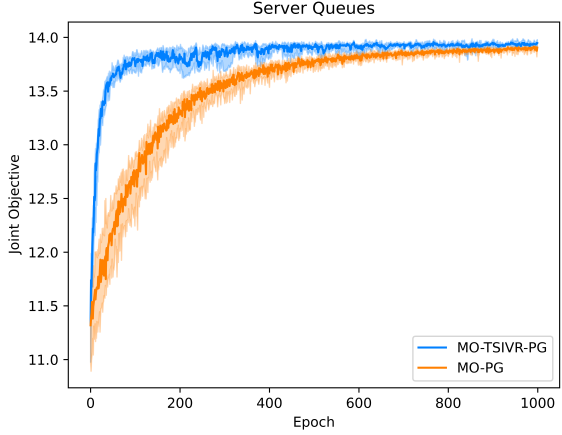

Figure 1: Algorithm 1 vs Algorithm 2 on the DeepSeaTreasure environment

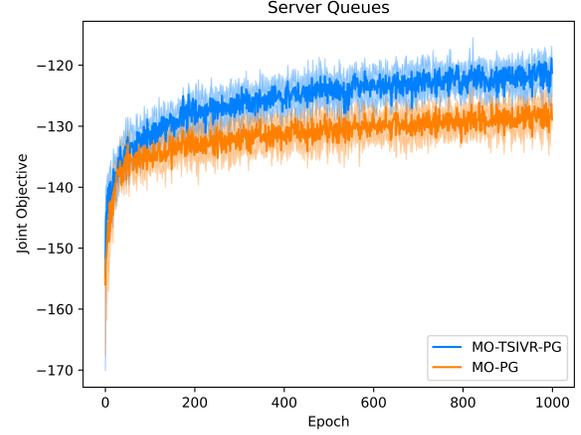

Figure 2: Algorithm 1 vs Algorithm 2 on the Server Queues environment with $M = 8$

## Global Convergence

We present two theorems bounding the sample complexity required to achieve global convergence. Notice that, as opposed to our analysis to establish stationary convergence, here we need to include Assumptions 4 and 5.

For Algorithm 1, we show the following result.

**Theorem 3**
Let Assumptions 1, 2, 3, 4 and 5 hold. Choosing $T = M^2\varepsilon^{-1}, N = M^4\varepsilon^{-2}, H = \frac{\ln(M\varepsilon^{-1})}{1-\gamma}$ and $\eta = \frac{1}{2L_\theta}$, Algorithm 1 achieves $\mathbb{E}\left[f(\mathbf{J}(\theta^*)) - f(\mathbf{J}(\tilde{\theta}_T))\right] \leq \tilde{\mathcal{O}}(\varepsilon)$ with sample complexity

$$T \times (2N) \times H = \tilde{\mathcal{O}}(\frac{M^6}{\varepsilon^3})$$

Following the same analysis, we obtain the following theoretical guarantees for Algorithm 2.

**Theorem 4**
Under the same assumptions of Theorem 3. Choosing $T = \ln(\varepsilon^{-1}), m = M^2\varepsilon^{-1}\ln(\varepsilon^{-1}), B = M^3\varepsilon^{-1}\ln(\varepsilon^{-1}), N = mB = M^5\varepsilon^{-2}(\ln(\varepsilon^{-1}))^2, H = \frac{\ln(M\varepsilon^{-1})}{1-\gamma}$ and $\eta = \frac{1}{2L_\theta + 8C_3/(ML_\theta)}$, Algorithm 2 achieves $\mathbb{E}\left[f(\mathbf{J}(\theta^*)) - f(\mathbf{J}(\theta_T))\right] \leq \tilde{\mathcal{O}}(\varepsilon)$ with sample complexity

$$T \times (2N + 2(m-1)B) \times H = \tilde{\mathcal{O}}(\frac{M^5}{\varepsilon^2}).$$

As for the results on stationary convergence, our algorithm obtains a better result compared to MO-PG for global convergence as well, improving the sample complexity by a factor $\frac{M}{\varepsilon}$.

Interestingly, notice that, for global convergence, both algorithms show a better dependency on $\varepsilon$, while the dependency on the number of objectives is worse by a factor of $M^2$.

## Experimental results

In this section, we aim to evaluate the performance of Algorithm 2 with two simple numerical experiments. The first one compares empirically MO-PG against MO-TSIVR-PG. The second aims to validate the theoretical sample complexity obtained in Theorem 4.

### Environments

We consider the following two environments and their respective scalarization:

1. **DeepSeaTreasure**: The DeepSeaTreasure environment is a tabular MDP with finite horizon and finite state and action spaces. The reward vector is 2-dimensional whose first component represents the value of the treasure at the current position, while the second component represents the time penalty (set to $-1$). Informally, the objective is to maximize the value of the treasure and minimize the time taken to reach it. Following previous literature (Agarwal, Aggarwal, and Lan (2022)), we consider the scalarization function

$$f(J_1, J_2) = \sqrt{J_1 + \sigma} + \sqrt{100 + J_2 + \sigma}$$

where the term $\sigma$ serves for numerical stability. We use $\sigma = 1$.

We use the public implementation of DeepSeaTreasure from MO-Gymnasium (Felten et al. 2023).

2. **Server Queues**: The Server Queues environment models a server managing $M$ queues, each with clients arriving according to Poisson processes with rates $\{\lambda_i\}_{i=1}^M$. At each discrete time-step, the server selects a queue to serve. Formally, this environment is modeled as an MDP with a finite horizon $H$, a state space of size $\binom{M+H-1}{M-1}$, and an action space of size $M$. The reward vector is $M$-dimensional, with all components set to zero except for the component corresponding to the served queue. The objective is to serve the queues as fairly as possible, using the same scalarization function ($\alpha$-fairness with $\alpha = 2$) presented in Bai, Agarwal, and Aggarwal (2022):

$$f(J_1, \ldots, J_M) = -\sum_{m=1}^{M} \frac{H}{J_m + \sigma}$$

where the term $\sigma$ serves for numerical stability. We use $\sigma = 1$.

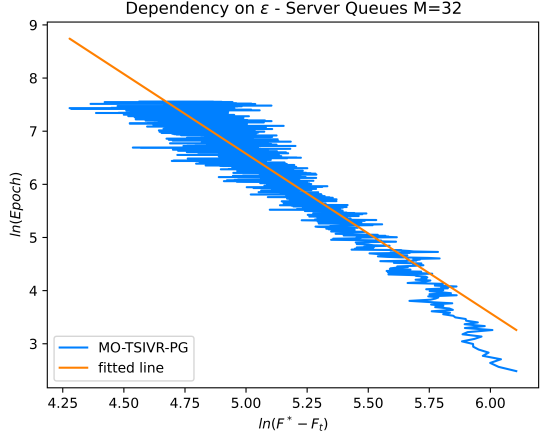

Figure 3: Log plot of the optimality gap against the number of epochs on the Server Queues environment with $M = 32$

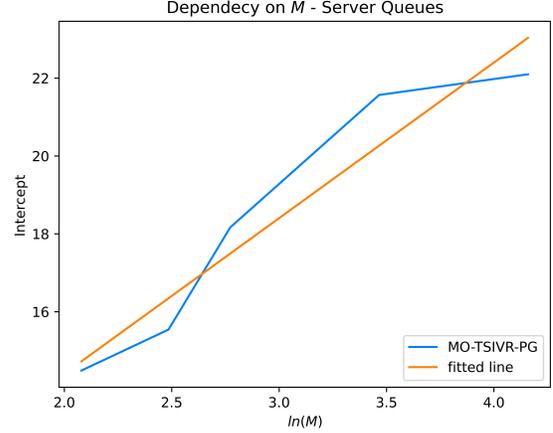

Figure 4: Plot of the dependency between $\ln(M)$ and $q_M$

This environment has already been considered for the setting of Multi-Objective RL in Agarwal, Aggarwal, and Lan (2022) and Bai, Agarwal, and Aggarwal (2022). However, to the best of our knowledge, no public implementation is available. Therefore, the implementation we use is original to this work.

### Comparison between Algorithms 1 and 2

We run both algorithms on DeepSeaTreasure with $\gamma = 1$ and Server Queues with $M = 8, H = 100, \gamma = 0.9999$. We set the parameters of Algorithm 1 as follows: $T = 1000, N = 288$. Moreover, we set the parameters of Algorithm 2 as follows: $T = 1000, N = 144, B = 12, m = 13$. Notice that the number of episodes per epoch, as well as the number of epochs, is the same for both algorithms, ensuring a fair comparison between the two. We execute 16 independent runs of each algorithm on both environments. We plot the median and the 0.25 and 0.75 confidence intervals.

As shown in figures 1 and 2, our algorithm outperforms the one from Bai, Agarwal, and Aggarwal (2022). This result indicates that our algorithm not only presents theoretical improvements but is also more effective in practice for the setting of Multiple-Objective Reinforcement Learning.

### Dependency on $\varepsilon$ and $M$ in practice

In the previous section, we provide theoretical bounds on the sample complexity of MO-TSIVR-PG. In this section, we aim to estimate the dependency of the sample complexity on the optimality gap $\varepsilon$ and the number of objectives $M$ in practice. Formally, let the sample complexity be of the order $\tilde{O}(\frac{M^a}{\varepsilon^b})$, our goal is to build estimators for $a$ and $b$.

For all $M \in \mathcal{M} = \{8, 12, 16, 32, 64\}$,

- We execute 8 independent runs of MO-TSIVR-PG, using the same parameters as in Section 5.2, on the Server Queues environment with $H = 500$ and $\gamma = 0.9999$ for 2400 epochs. We estimate the optimal value $f_M^*$ as the maximum performance achieved over all the timesteps of each run. Moreover, we also estimate the optimality gap after $t$ epochs $\varepsilon_t^M$ as the gap between the median value $f_t^M$ achieved at every timestep and $f_M^*$.

- Next, we approximate the relation between the number of epochs passed and the optimality gap as

$$\ln(T) \approx q_M - b_M \ln(\varepsilon_T)$$

and we estimate $q_M$ and $b_M$ with linear interpolation as shown in Figure 3.

We conclude by estimating $b$ as $\hat{b} = \frac{1}{|\mathcal{M}|} \sum_{M \in \mathcal{M}} b_M$. Moreover, we estimate $a$ as the slope of the line obtained by interpolation on $\{(\ln(M), q_M)\}_{M \in \mathcal{M}}$ as shown in Figure 4.

The final results indicate a sample complexity approximately in the order of $\tilde{O}(\frac{M^4}{\varepsilon^3})$. The dependency on the optimality gap $\varepsilon$ matches almost exactly theoretical guarantees from Theorem 2. Meanwhile, the dependency on the number of objectives $M$ is in the middle between the guarantees provided by 2 and 4.

### Conclusion

The Multi-Objective Reinforcement Learning (MORL) framework, while less explored than traditional Reinforcement Learning, is relevant for numerous real-world applications. This work builds upon the results from Bai, Agarwal, and Aggarwal (2022) and we propose a novel algorithm, improving over both theoretical guarantees and experimental results. Compared to other algorithms leveraging variance reduction (Zhang et al. (2021), Barakat, Fatkhullin, and He

(2023)), our algorithm can operate over large and continuous state-action spaces.

Notice that we operate in a setting where a scalarization function to balance the objectives is already fixed. Future research directions include investigating algorithms that do not rely on this assumption and exploring the Pareto optimal frontier to bridge the gap between this area of Multi-Objective Reinforcement Learning and cooperative Multi-Agent Reinforcement Learning.

# Appendix

## A Proof of Lemma 3

**Lemma A.1**
For any policies $\pi_{\theta_1}, \pi_{\theta_2}$, let Assumption 2 hold, the importance weight satisfies
$$\mathbb{V}[\omega_t(\tau|\theta_1, \theta_2)] \leq \exp\{2G(t+1)\|\theta_1 - \theta_2\|\}, \forall t \in \{0, \ldots, H-1\}.$$

*Proof of Lemma A.1.* For all $(s, a) \in \mathcal{S} \times \mathcal{A}$, note that
$$\frac{\pi_{\theta_2}(a|s)}{\pi_{\theta_1}(a|s)} = \exp\{\log \pi_{\theta_2}(a|s) - \log \pi_{\theta_1}(a|s)\}$$
$$\leq \exp\{G\|\theta_2 - \theta_1\|\}. \quad \text{(by Assumption 2)}$$

Therefore, we have
$$\omega_t(\tau|\theta_1, \theta_2) = \prod_{t'=0}^{t} \frac{\pi_{\theta_2}(a_{t'}|s_{t'})}{\pi_{\theta_1}(a_{t'}|s_{t'})}$$
$$\leq \exp\{G(t+1)\|\theta_2 - \theta_1\|\}.$$

Finally, we conclude that
$$\mathbb{V}[\omega_t(\tau|\theta_1, \theta_2)] \leq \exp\{2G(t+1)\|\theta_2 - \theta_1\|\}$$
since $\mathbb{V}[X] \leq \mathbb{E}[X^2]$. □

*Proof of Lemma 3.* Thanks to the upper bound on the variance of the importance sampling weights given by Lemma A.1 and using Assumption 2, Lemma 3 simply comes from Lemma B.1 of [7] □

## B Auxiliary Lemmas

**Lemma B.1**
Let $\tau = \{s_0, a_0, \ldots, s_{H-1}, a_{H-1}\}$ be an arbitrary trajectories. Then, we have the following upper bounds:

(i) For all $\theta$ and for all $J', J'' \in \Omega$, it holds that $\|g(\tau|\theta, J') - g(\tau|\theta, J'')\| \leq \frac{MGL_f}{(1-\gamma)^2}\|J' - J''\|$.

(ii) For all $\theta_1, \theta_2$ and for all $J \in \Omega$, it holds that $\|g(\tau|\theta_1, J) - g(\tau|\theta_2, J)\| \leq \frac{MCS}{(1-\gamma)^2}\|\theta_1 - \theta_2\|$

(iii) For all $\theta$, it holds that $\|\nabla_\theta \mathbf{J}^H(\theta)\| \leq \frac{G\sqrt{M}}{(1-\gamma)^2}$

(iv) For all $\theta$, it holds that $\|\nabla_\theta f(\mathbf{J}^H(\theta))\| \leq \frac{CGM}{(1-\gamma)^2}$

*Proof.* (i) For the first claim, we have
$$\|g(\tau|\theta, \mathbf{J}') - g(\tau|\theta, \mathbf{J}'')\|$$
$$= \left\| \sum_{m=1}^{M} \left( \frac{\partial f}{\partial J_m}(\mathbf{J}') - \frac{\partial f}{\partial J_m}(\mathbf{J}'') \right) \sum_{t=0}^{H-1} \nabla_\theta \log \pi_\theta(a_t|s_t) \sum_{h=t}^{H-1} \gamma^h r_m(s_h, a_h) \right\|$$
$$\leq \sum_{m=1}^{M} \left| \frac{\partial f}{\partial J_m}(\mathbf{J}') - \frac{\partial f}{\partial J_m}(\mathbf{J}'') \right| \sum_{t=0}^{H-1} \|\nabla_\theta \log \pi_\theta(a_t|s_t)\| \sum_{h=t}^{H-1} \gamma^h |r_m(s_h, a_h)|$$
$$\leq L_f \|\mathbf{J}' - \mathbf{J}''\| \sum_{m=1}^{M} \sum_{t=0}^{H-1} \|\nabla_\theta \log \pi_\theta(a_t|s_t)\| \sum_{h=t}^{H-1} \gamma^h |r_m(s_h, a_h)| \quad \text{(by Assumption 3)}$$
$$\leq GL_f \|\mathbf{J}' - \mathbf{J}''\| \sum_{m=1}^{M} \sum_{t=0}^{H-1} \sum_{h=t}^{H-1} \gamma^h |r_m(s_h, a_h)| \quad \text{(by Assumption 2)}$$
$$\leq \frac{MGL_f}{(1-\gamma)^2} \|\mathbf{J}' - \mathbf{J}''\|. \quad \text{(by Assumption 1)}$$

(ii) For the second claim, we have

$$\|g(\tau|\theta_1, \mathbf{J}) - g(\tau|\theta_2, \mathbf{J})\|$$

$$=\|\sum_{m=1}^{M} \frac{\partial f}{\partial J_m}(\mathbf{J}) (\boldsymbol{\nabla}_\theta \log \pi_{\theta_1}(a_t|s_t) - \boldsymbol{\nabla}_\theta \log \pi_{\theta_2}(a_t|s_t)) \sum_{h=t}^{H-1} \gamma^h r_m(s_h, a_h)\|$$

$$\leq \sum_{m=1}^{M} \left|\frac{\partial f}{\partial J_m}(\mathbf{J})\right| \sum_{t=0}^{H-1} \|\boldsymbol{\nabla}_\theta \log \pi_{\theta_1}(a_t|s_t) - \boldsymbol{\nabla}_\theta \log \pi_{\theta_2}(a_t|s_t)\| \sum_{h=t}^{H-1} \gamma^h |r_m(s_h, a_h)|$$

$$\leq C \sum_{m=1}^{M} \sum_{t=0}^{H-1} \|\boldsymbol{\nabla}_\theta \log \pi_{\theta_1}(a_t|s_t) - \boldsymbol{\nabla}_\theta \log \pi_{\theta_2}(a_t|s_t)\| \sum_{h=t}^{H-1} \gamma^h |r_m(s_h, a_h)| \quad \text{(by Lemma 1)}$$

$$\leq CS\|\theta_1 - \theta_2\| \sum_{m=1}^{M} \sum_{t=0}^{H-1} \sum_{h=t}^{H-1} \gamma^h |r_m(s_h, a_h)| \quad \text{(by Assumption 2)}$$

$$\leq \frac{MCS}{(1-\gamma)^2} \|\theta_1 - \theta_2\|. \quad \text{(by Assumption 1)}$$

(iii) For the third claim, we have

$$\|\boldsymbol{\nabla}_\theta \mathbf{J}^H(\theta)\|$$

$$= \|\mathbb{E}\left[\sum_{t=0}^{H-1} \boldsymbol{\nabla}_\theta \log \pi_\theta(a_t|s_t) \sum_{h=t}^{H-1} \gamma^h \mathbf{r}(s_h, a_h)\right]\|$$

$$\leq \mathbb{E}\left[\sum_{t=0}^{H-1} \|\boldsymbol{\nabla}_\theta \log \pi_\theta(a_t|s_t)\| \sum_{h=t}^{H-1} \gamma^h \|\mathbf{r}(s_h, a_h)\|\right]$$

$$\leq \frac{G\sqrt{M}}{(1-\gamma)^2}. \quad \text{(by Assumptions 1 and 2)}$$

(iv) For the fourth claim, we have

$$\|\nabla_\theta f(\mathbf{J}^H(\theta))\| \leq \|\boldsymbol{\nabla}_\theta \mathbf{J}^H(\theta)\| \cdot \|\nabla_\mathbf{J} f(\mathbf{J}^H(\theta))\|$$

$$\leq \frac{CGM}{(1-\gamma)^2}. \quad \text{(by Assumption 3 and part (iii))}$$

□

**Lemma B.2**
Let $\tau = \{s_0, a_0, \ldots, s_{H-1}, a_{H-1}\}$ denote a truncated trajectory sample according $\pi_{\theta_1}$. Then, it holds that

$$\mathbb{E}_{\tau \sim \pi_{\theta_1}}[\mathbf{J}(\tau|\theta_1, \theta_2)] = \mathbf{J}^H(\theta_2) \quad \text{and} \quad \mathbb{E}_{\tau \sim \pi_{\theta_1}}\left[g(\tau|\theta_1, \theta_2, \hat{\mathbf{J}})\right] = \left(\boldsymbol{\nabla}_\theta \mathbf{J}^H(\theta_2)\right)^\top \boldsymbol{\nabla}_\mathbf{J} f(\hat{\mathbf{J}})$$

for any $\theta_1, \theta_2$ and for all $\hat{\mathbf{J}} \in \Omega$.

*Proof.* The proof is extremely similar to the proof of Proposition E.1 in Zhang et al. [8]. □

**Lemma B.3**

$$\mathbb{E}[\|\mathcal{G}_\eta(\theta)\|] \leq \varepsilon \implies \mathbb{E}[\|\nabla_\theta f(\mathbf{J}(\theta))\|] \leq \left(1 + \frac{\eta}{\delta} \cdot \frac{CGM}{(1-\gamma)^2}\right) \cdot \varepsilon$$

*Proof.* Let $\mathcal{E}$ denote the event $\eta\|\nabla_\theta f(\mathbf{J}(\theta))\| < \delta$. Then, we have

$$\varepsilon \geq \mathbb{E}[\|\mathcal{G}_\eta(\theta)\|]$$
$$= \Pr[\mathcal{E}] \cdot \mathbb{E}[\|\mathcal{G}_\eta(\theta)\| | \mathcal{E}] + \Pr[\mathcal{E}^c] \cdot \mathbb{E}[\|\mathcal{G}_\eta(\theta)\| | \mathcal{E}^c]$$
$$= \Pr[\mathcal{E}] \cdot \mathbb{E}[\|\nabla_\theta f(\mathbf{J}(\theta))\| | \mathcal{E}] + \Pr[\mathcal{E}^c] \cdot \frac{\delta}{\eta}.$$

Therefore, we have

$$\Pr[\mathcal{E}] \cdot \mathbb{E}\left[\|\boldsymbol{\nabla}_\theta f(\mathbf{J}(\theta))\| | \mathcal{E}\right] \leq \varepsilon \quad \text{and} \quad \Pr[\mathcal{E}^c] \leq \frac{\eta\varepsilon}{\delta}.$$

Thus, it holds that

$$\mathbb{E}\left[\|\boldsymbol{\nabla}_\theta f(\mathbf{J}(\theta))\|\right] \leq \Pr[\mathcal{E}] \cdot \mathbb{E}\left[\|\boldsymbol{\nabla}_\theta f(\mathbf{J}(\theta))\| | \mathcal{E}\right] + \Pr[\mathcal{E}^c] \cdot \mathbb{E}\left[\|\boldsymbol{\nabla}_\theta f(\mathbf{J}(\theta))\| | \mathcal{E}^c\right]$$
$$\leq \varepsilon + \frac{\eta\varepsilon}{\delta} \cdot \frac{CGM}{(1-\gamma)^2}. \qquad \text{(by Lemma B.1)}$$

□

## C  Bounding the Variance

### C.1  Proof of Lemma 4

#### C.1.1  Variance of $J_i$

**Lemma C.1**

$$\mathbb{E}\left[\|J_i - \mathbf{J}^H(\theta_i)\|^2\right] \leq \frac{M}{(1-\gamma)^2 N}$$

*Proof.*

$$\mathbb{E}[\|J_i - \mathbf{J}^H(\theta_i)\|^2]$$
$$= \mathbb{E}\left[\|\frac{1}{N}\sum_{\tau \in \mathcal{N}_1^i} \mathbf{J}(\tau|\theta_i) - \mathbf{J}^H(\theta_i)\|^2\right] \qquad \text{(by Algorithm 1)}$$
$$= \frac{1}{N} \cdot \mathbb{E}\left[\|\mathbf{J}(\tau|\theta_i) - \mathbf{J}^H(\theta_i)\|^2\right] \qquad \text{(by indep of the trajectories and Lemma B.2)}$$
$$\leq \frac{1}{N} \cdot \mathbb{E}\left[\|\mathbf{J}(\tau|\theta_i)\|^2\right] \qquad \text{(since } \mathbb{V}(X) \leq \mathbb{E}\left[\|X\|^2\right]\text{)}$$
$$\leq \frac{M}{(1-\gamma)^2 N} \qquad \text{(by Assumption 1)}$$

□

#### C.1.2  Supporting Lemma

**Lemma C.2**

$$\mathbb{E}\left[\|g_i - \left(\boldsymbol{\nabla}_\theta \mathbf{J}^H(\theta_i)\right)^\top \boldsymbol{\nabla}_\mathbf{J} f(J_i)\|^2\right] \leq \frac{6M^3 G^2 L_f^2}{N(1-\gamma)^6} + \frac{3M^2 C^2 G^2}{N(1-\gamma)^4}$$

*Proof.* Let us define $J_i^* = \mathbf{J}^H(\theta_i)$. Then

$$\mathbb{E}\left[\|g_i - \left(\boldsymbol{\nabla}_\theta \mathbf{J}^H(\theta_i)\right)^\top \boldsymbol{\nabla}_\mathbf{J} f(J_i)\|^2\right]$$
$$\leq 3\mathbb{E}\left[\|\frac{1}{N}\sum_{\tau \in \mathcal{N}_2^i}(g(\tau|\theta_i, J_i) - g(\tau|\theta_i, J_i^*))\|^2\right] + 3\mathbb{E}\left[\|\frac{1}{N}\sum_{\tau \in \mathcal{N}_2^i} g(\tau|\theta_i, J_i^*) - \boldsymbol{\nabla}_\theta f(\mathbf{J}^H(\theta_i))\|^2\right]$$
$$+ 3\mathbb{E}\left[\|\boldsymbol{\nabla}_\theta f(\mathbf{J}^H(\theta_i)) - \left(\boldsymbol{\nabla}_\theta \mathbf{J}^H(\theta_i)\right)^\top \boldsymbol{\nabla}_\mathbf{J} f(J_i)\|^2\right].$$

For the first term, we have

$$\mathbb{E}\left[\|\frac{1}{N}\sum_{\tau \in \mathcal{N}_2^i}(g(\tau|\theta_i, J_i) - g(\tau|\theta_i, J_i^*))\|^2\right]$$

$$\leq \frac{1}{N}\sum_{\tau \in \mathcal{N}_2^i} \mathbb{E}\left[\|g(\tau|\theta_i, J_i) - g(\tau|\theta_i, J_i^*)\|^2\right]$$

$$\leq \frac{M^2 G^2 L_f^2}{(1-\gamma)^4} \mathbb{E}\left[\|J_i - J_i^*\|^2\right] \qquad \text{(by lemma B.1)}$$

$$\leq \frac{M^3 G^2 L_f^2}{N(1-\gamma)^6} \qquad \text{(by lemma C.1)}$$

For the second term, we have

$$\mathbb{E}\left[\|\frac{1}{N}\sum_{\tau \in \mathcal{N}_2^i} g(\tau|\theta_i, J_i^*) - \boldsymbol{\nabla}_\theta f(\mathbf{J}^H(\theta_i))\|^2\right]$$

$$= \frac{1}{N} \mathbb{E}\left[\|g(\tau|\theta_i, J_i^*) - \boldsymbol{\nabla}_\theta f(\mathbf{J}^H(\theta_i))\|^2\right] \qquad \text{(by Lemma B.2)}$$

$$\leq \frac{1}{N} \mathbb{E}\left[\|g(\tau|\theta_i, J_i^*)\|^2\right]$$

$$= \frac{1}{N} \mathbb{E}\left[\|\sum_{m=1}^{M} \frac{\partial f}{\partial J_m}(J_i^*) \sum_{t=0}^{H-1} \boldsymbol{\nabla}_\theta \log \pi_{\theta_i}(a_t|s_t) \sum_{h=t}^{H-1} \gamma^h r_m(s_h, a_h)\|^2\right]$$

$$\leq \frac{M^2 C^2 G^2}{N(1-\gamma)^4}.$$

Finally, for the third term, we have

$$\mathbb{E}\left[\|\boldsymbol{\nabla}_\theta f(\mathbf{J}^H(\theta_i)) - \left(\boldsymbol{\nabla}_\theta \mathbf{J}^H(\theta_i)\right)^\top \boldsymbol{\nabla}_\mathbf{J} f(J_i)\|^2\right]$$

$$= \mathbb{E}\left[\|\left(\boldsymbol{\nabla}_\theta \mathbf{J}^H(\theta_i)\right)^\top (\boldsymbol{\nabla}_\mathbf{J} f(J_i^*) - \boldsymbol{\nabla}_\mathbf{J} f(J_i))\|^2\right]$$

$$\leq \mathbb{E}\left[\|\boldsymbol{\nabla}_\theta \mathbf{J}^H(\theta_i)\|^2 \|\boldsymbol{\nabla}_\mathbf{J} f(J_i^*) - \boldsymbol{\nabla}_\mathbf{J} f(J_i)\|^2\right]$$

$$\leq \frac{MG^2}{(1-\gamma)^4} \mathbb{E}\left[\|\boldsymbol{\nabla}_\mathbf{J} f(J_i^*) - \boldsymbol{\nabla}_\mathbf{J} f(J_i)\|^2\right] \qquad \text{(by Lemma B.1)}$$

$$\leq \frac{M^3 G^2 L_f^2}{(1-\gamma)^6 N}. \qquad \text{(by Assumption 3 and Lemma C.1)}$$

$\square$

### C.1.3 End of Proof

*Proof of Lemma 4.* First, we decompose the variance of the gradient estimates as follows:

$$\mathbb{E}\left[\|g_i - \boldsymbol{\nabla}_\theta f(\mathbf{J}(\theta_i))\|^2\right]$$
$$\leq 3\mathbb{E}\left[\|g_i - \boldsymbol{\nabla}_\theta \mathbf{J}^H(\theta_i)^\top \boldsymbol{\nabla}_\mathbf{J} f(J_i)\|^2\right] + 3\mathbb{E}\left[\|\boldsymbol{\nabla}_\theta \mathbf{J}^H(\theta_j^i)^\top \boldsymbol{\nabla}_\mathbf{J} f(J_i) - \boldsymbol{\nabla}_\theta f(\mathbf{J}^H(\theta_i))\|^2\right]$$
$$+ 3\mathbb{E}\left[\boldsymbol{\nabla}_\theta f(\mathbf{J}^H(\theta_i)) - \boldsymbol{\nabla}_\theta f(\mathbf{J}(\theta_i))\|^2\right].$$

Note that Lemma C.2 gives an upper bound on the first term. Next, the second term is upper-bounded as follows:

$$\mathbb{E}\left[\|\nabla_\theta \mathbf{J}^H(\theta_i)^\top \left(\nabla_\mathbf{J} f(J_i) - \nabla_\mathbf{J} f(\mathbf{J}^H(\theta_i))\right)\|^2\right]$$

$$\leq \frac{MG^2}{(1-\gamma)^4} \cdot \mathbb{E}\left[\|\nabla_\mathbf{J} f(J_i) - \nabla_\mathbf{J} f(\mathbf{J}^H(\theta_i))\|^2\right] \qquad \text{(by Lemma B.1)}$$

$$\leq \frac{M^2 G^2 L_f^2}{(1-\gamma)^4} \cdot \mathbb{E}\left[\|J_i - \mathbf{J}^H(\theta_i)\|^2\right] \qquad \text{(by Assumption 3)}$$

$$\leq \frac{M^2 G^2 L_f^2}{(1-\gamma)^4} \cdot \left(\frac{M}{(1-\gamma)^2 N}\right) \qquad \text{(by lemma C.1)}$$

The last term is upper bounded directly by lemma 2.
Combining the inequalities we obtain

$$\mathbb{E}\left[\|g_j^i - \nabla_\theta f(\mathbf{J}(\theta_i))\|^2\right] \leq \frac{C_1}{N} + C_2 \gamma^{2H}$$

where

$$C_1 = \frac{21 M^3 G^2 L_f^2}{(1-\gamma^6)} + \frac{9 M^2 C^2 G^2}{(1-\gamma)^4},$$

$$C_2 = \frac{3 M^2 G^2}{(1-\gamma)^4} \left[\sqrt{M} L_f \frac{1 - \gamma^H - H\gamma^H(1-\gamma)}{1-\gamma} + C[1 + H(1-\gamma)]\right]^2.$$

□

## C.2 Proof of Lemma 5

### C.2.1 Variance of $J_j^i$

To prove Lemma 5, we bound the variance of the cumulative rewards estimator $J_j^i$ with the following supporting lemma.

**Lemma C.3**
For the cumulative rewards estimators $J_j^i$, we have

$$\mathbb{E}\left[\|J_j^i - \mathbf{J}^H(\theta_j^i)\|^2\right] \leq \frac{M}{(1-\gamma)^2 N} + \frac{2MH(2G^2 + S)(\exp\{2GH\delta\} + 1)}{(1-\gamma)^3 B} \cdot \sum_{j'=1}^{j} \mathbb{E}\left[\|\theta_{j-1}^i - \theta_j^i\|^2\right]$$

*Proof.* We proceed with a proof by induction on $j$. First, let us consider the base case of $j = 0$.

$$\mathbb{E}[\|J_0^i - \mathbf{J}^H(\theta_0^i)\|^2]$$

$$= \mathbb{E}\left[\|\frac{1}{N} \sum_{\tau \in \mathcal{N}_1^i} \mathbf{J}(\tau|\theta_0^i) - \mathbf{J}^H(\theta_0^i)\|^2\right] \qquad \text{(by Algorithm 2)}$$

$$= \frac{1}{N} \cdot \mathbb{E}\left[\|\mathbf{J}(\tau|\theta_0^i) - \mathbf{J}^H(\theta_0^i)\|^2\right] \qquad \text{(by indep of the trajectories and Lemma B.2)}$$

$$\leq \frac{1}{N} \cdot \mathbb{E}\left[\|\mathbf{J}(\tau|\theta_0^i)\|^2\right] \qquad \text{(since } \mathbb{V}(X) \leq \mathbb{E}\left[\|X\|^2\right]\text{)}$$

$$\leq \frac{M}{(1-\gamma)^2 N} \qquad \text{(by Assumption 1)}$$

Now, for the inductive step, let the lemma hold for $j-1$ for some $j \geq 1$, then we have

$$\mathbb{E}\left[\|J_j^i - \mathbf{J}^H(\theta_j^i)\|^2\right]$$

$$= \mathbb{E}\left[\|\frac{1}{B}\sum_{\tau \in \mathcal{B}_{j,1}^i}\left(\mathbf{J}(\tau|\theta_j^i) - \mathbf{J}_\omega(\tau|\theta_j^i, \theta_{j-1}^i)\right) + J_{j-1}^i - \mathbf{J}^H(\theta_j^i)\|^2\right] \quad \text{(by definition of } J_j^i\text{)}$$

$$= \mathbb{E}\left[\|\frac{1}{B}\sum_{\tau \in \mathcal{B}_{j,1}^i}\left(\mathbf{J}(\tau|\theta_j^i) - \mathbf{J}_\omega(\tau|\theta_j^i, \theta_{j-1}^i)\right) + J_{j-1}^i - \mathbf{J}^H(\theta_{j-1}^i) + \mathbf{J}^H(\theta_{j-1}^i) - \mathbf{J}^H(\theta_j^i)\|^2\right]$$

$$= \mathbb{E}\left[\|\frac{1}{B}\sum_{\tau \in \mathcal{B}_{j,1}^i}\left(\mathbf{J}(\tau|\theta_j^i) - \mathbf{J}_\omega(\tau|\theta_j^i, \theta_{j-1}^i)\right) + \mathbf{J}^H(\theta_{j-1}^i) - \mathbf{J}^H(\theta_j^i)\|^2\right]$$

$$+ \mathbb{E}\left[\|J_{j-1}^i - \mathbf{J}^H(\theta_{j-1}^i)\|^2\right]$$

$$+ \underbrace{2\mathbb{E}\left[\langle\frac{1}{B}\sum_{\tau \in \mathcal{B}_{j,1}^i}\left(\mathbf{J}(\tau|\theta_j^i) - \mathbf{J}_\omega(\tau|\theta_j^i, \theta_{j-1}^i)\right) + \mathbf{J}^H(\theta_{j-1}^i) - \mathbf{J}^H(\theta_j^i), J_{j-1}^i - \mathbf{J}^H(\theta_{j-1}^i)\rangle\right]}_{=0} \quad (1)$$

where the cross term is zero in 1 since

$$\mathbb{E}\left[\frac{1}{B}\sum_{\tau \in \mathcal{B}_{j,1}^i}\mathbf{J}(\tau|\theta_j^i) - \mathbf{J}^H(\theta_j^i)\right] = \mathbb{E}\left[\frac{1}{B}\sum_{\tau \in \mathcal{B}_{j,1}^i}\mathbf{J}_\omega(\tau|\theta_j^i, \theta_{j-1}^i) - \mathbf{J}^H(\theta_{j-1}^i)\right] = 0$$

and since $J_{j-1}^i - \mathbf{J}^H(\theta_{j-1}^i)$ does not depend on the trajectories in $\mathcal{B}_{j,1}^i$. Moreover, the first term can be bounded as follows:

$$\mathbb{E}\left[\|\frac{1}{B}\sum_{\tau \in \mathcal{B}_{j,1}^i}\left(\mathbf{J}(\tau|\theta_j^i) - \mathbf{J}_\omega(\tau|\theta_j^i, \theta_{j-1}^i)\right) + \mathbf{J}^H(\theta_{j-1}^i) - \mathbf{J}^H(\theta_j^i)\|^2\right]$$

$$= \frac{1}{B}\mathbb{E}\left[\|\mathbf{J}(\tau|\theta_j^i) - \mathbf{J}_\omega(\tau|\theta_j^i, \theta_{j-1}^i) + \mathbf{J}^H(\theta_{j-1}^i) - \mathbf{J}^H(\theta_j^i)\|^2\right] \quad \text{(by Lemma B.2)}$$

$$= \frac{1}{B}\mathbb{E}\left[\|\mathbf{J}(\tau|\theta_j^i) - \mathbf{J}_\omega(\tau|\theta_j^i, \theta_{j-1}^i)\|^2\right] \quad \text{(since } \mathbb{V}(X) \leq \mathbb{E}\left[\|X\|^2\right]\text{)}$$

$$= \frac{1}{B}\mathbb{E}\left[\|\sum_{t=0}^{H-1}\gamma^t(1 - \omega_t(\tau|\theta_j^i, \theta_{j-1}^i))\mathbf{r}(s_t, a_t)\|^2\right] \quad \text{(by definition of } \mathbf{J}(\tau|\theta_j^i) \text{ and } \mathbf{J}_\omega(\tau|\theta_j^i, \theta_{j-1}^i)\text{)}$$

$$\leq \frac{H}{B}\mathbb{E}\left[\sum_{t=0}^{H-1}\gamma^{2t}|1 - \omega_t(\tau|\theta_j^i, \theta_{j-1}^i)|^2\|\mathbf{r}(s_t, a_t)\|^2\right] \quad \text{(by AM-QM)}$$

$$\leq \frac{MH}{B}\sum_{t=0}^{H-1}\gamma^{2t}\mathbb{V}(\omega_t(\tau|\theta_j^i, \theta_{j-1}^i)) \quad \text{(since } \mathbb{E}\left[\omega_t(\tau|\theta_j^i, \theta_{j-1}^i)\right] = 1\text{)}$$

$$\leq \frac{MH \cdot \mathbb{E}\left[\|\theta_{j-1}^i - \theta_j^i\|^2\right]}{B}\sum_{t=0}^{H-1}\gamma^{2t}t(2tG^2 + S)(\exp\{2GH\delta\} + 1) \quad \text{(by Lemma 3)}$$

$$\leq \frac{MH \cdot \mathbb{E}\left[\|\theta_{j-1}^i - \theta_j^i\|^2\right]}{B}\sum_{t=0}^{H-1}\gamma^{2t}t(t+1)(2G^2 + S)(\exp\{2GH\delta\} + 1)$$

$$\leq \frac{2MH(2G^2 + S)(\exp\{2GH\delta\} + 1)}{(1-\gamma^2)^3 B}\mathbb{E}\left[\|\theta_{j-1}^i - \theta_j^i\|^2\right]$$

$$\leq \frac{2MH(2G^2 + S)(\exp\{2GH\delta\} + 1)}{(1-\gamma)^3 B}\mathbb{E}\left[\|\theta_{j-1}^i - \theta_j^i\|^2\right].$$

Plugging this results back in 1, we have

$$\mathbb{E}\left[\|J_j^i - \mathbf{J}^H(\theta_j^i)\|^2\right]$$
$$\leq \mathbb{E}\left[J_{j-1}^i - \mathbf{J}^H(\theta_{j-1}^i)\|^2\right] + \frac{2MH(2G^2+S)(\exp\{2GH\delta\}+1)}{(1-\gamma)^3 B}\mathbb{E}\left[\|\theta_{j-1}^i - \theta_j^i\|^2\right]$$
$$\leq \frac{M}{(1-\gamma)^2 N} + \frac{2MH(2G^2+S)(\exp\{2GH\delta\}+1)}{(1-\gamma)^3 B} \cdot \sum_{j'=1}^{j}\mathbb{E}\left[\|\theta_{j-1}^i - \theta_j^i\|^2\right]. \quad \text{(by induction)}$$

$\square$

**Corollary 1**

$$\mathbb{E}\left[\|\boldsymbol{\nabla}_\mathbf{J} f(P_0^i) - \boldsymbol{\nabla}_\mathbf{J} f(\mathbf{J}^H(\theta_0^i))\|^2\right] \leq \frac{M^2 L_f^2}{(1-\gamma)^2 N}.$$

*Proof.*

$$\mathbb{E}\left[\|\boldsymbol{\nabla}_\mathbf{J} f(P_0^i) - \boldsymbol{\nabla}_\mathbf{J} f(\mathbf{J}^H(\theta_0^i))\|^2\right]$$
$$\leq ML_f^2 \mathbb{E}\left[\|P_0^i - \mathbf{J}^H(\theta_0^i)\|^2\right] \quad \text{(by Assumption 3)}$$
$$\leq ML_f^2 \mathbb{E}\left[\|J_0^i - \mathbf{J}^H(\theta_0^i)\|^2\right] \quad \text{(by non-expansiveness of } \mathbf{Proj}\text{)}$$
$$\leq \frac{M^2 L_f^2}{(1-\gamma)^2 N} \quad \text{(by Lemma C.3 for } j=0\text{)}$$

$\square$

### C.2.2 Two Supporting Lemmas
**Lemma C.4**

$$\mathbb{E}\left[\|g_j^i - \left(\boldsymbol{\nabla}_\theta \mathbf{J}^H(\theta_j^i)\right)^\top \boldsymbol{\nabla}_\mathbf{J} f(P_j^i))\|^2\right] - \mathbb{E}\left[\|g_0^i - \left(\boldsymbol{\nabla}_\theta \mathbf{J}^H(\theta_0^i)\right)^\top \boldsymbol{\nabla}_\mathbf{J} f(P_j^i)\|^2\right]$$
$$\leq \left(\frac{3M^2 C^2 S^2}{(1-\gamma)^4} + \frac{6M^2 G^2 H(2G^2+S)(\exp\{2GH\delta\}+1)}{(1-\gamma)^5}\left(12C^2 + \frac{ML_f^2}{(1-\gamma)^2}\right)\right) \cdot \frac{\sum_{j'=1}^{j}\mathbb{E}\left[\|\theta_{j'-1}^i - \theta_{j'}^i\|^2\right]}{B}$$

*Proof.* Trajectories $\tau \in \mathcal{B}_{j,2}^i$ sampled under policy $\pi_{\theta_j^i}$:

$$\mathbb{E}\left[\|g_j^i - \left(\boldsymbol{\nabla}_\theta \mathbf{J}^H(\theta_j^i)\right)^\top \boldsymbol{\nabla}_\mathbf{J} f(P_j^i)\|^2\right] - \mathbb{E}\left[\|g_{j-1}^i - \left(\boldsymbol{\nabla}_\theta \mathbf{J}^H(\theta_{j-1}^i)\right)^\top \boldsymbol{\nabla}_\mathbf{J} f(P_{j-1}^i)\|^2\right]$$
$$= \mathbb{E}\left[\|\frac{1}{B}\sum_{\tau \in \mathcal{B}_{j,1}^i}\left(g(\tau|\theta_j^i, P_j^i) - g(\tau|\theta_j^i, \theta_{j-1}^i, P_{j-1}^i)\right) + g_{j-1}^i - \left(\boldsymbol{\nabla}_\theta \mathbf{J}^H(\theta_{j-1}^i)\right)^\top \boldsymbol{\nabla}_\mathbf{J} f(P_{j-1}^i)\right.$$
$$\left. + \left(\boldsymbol{\nabla}_\theta \mathbf{J}^H(\theta_{j-1}^i)\right)^\top \boldsymbol{\nabla}_\mathbf{J} f(P_{j-1}^i) - \left(\boldsymbol{\nabla}_\theta \mathbf{J}^H(\theta_j^i)\right)^\top \boldsymbol{\nabla}_\mathbf{J} f(P_j^i)\|^2\right]$$
$$- \mathbb{E}\left[\|g_{j-1}^i - \left(\boldsymbol{\nabla}_\theta \mathbf{J}^H(\theta_{j-1}^i)\right)^\top \boldsymbol{\nabla}_\mathbf{J} f(P_{j-1}^i)\|^2\right]$$
$$= \mathbb{E}\left[\|\frac{1}{B}\sum_{\tau \in \mathcal{B}_{j,1}^i}\left(g(\tau|\theta_j^i, P_j^i) - g(\tau|\theta_j^i, \theta_{j-1}^i, P_{j-1}^i)\right) + \left(\boldsymbol{\nabla}_\theta \mathbf{J}^H(\theta_{j-1}^i)\right)^\top \boldsymbol{\nabla}_\mathbf{J} f(P_{j-1}^i)\right.$$
$$\left. - \left(\boldsymbol{\nabla}_\theta \mathbf{J}^H(\theta_j^i)\right)^\top \boldsymbol{\nabla}_\mathbf{J} f(P_j^i)\|^2\right] + \text{(cross-term)}.$$

Note that the cross-term above is zero since

$$\mathbb{E}\left[\frac{1}{B}\sum_{\tau\in\mathcal{B}_{j,1}^i} g(\tau|\theta_j^i, P_j^i)\bigg|\mathcal{F}_j^i\right] = \left(\boldsymbol{\nabla}_\theta \mathbf{J}^H(\theta_j^i)\right)^\top \boldsymbol{\nabla}_{\mathbf{J}} f(P_j^i) \quad \text{and}$$

$$\mathbb{E}\left[\frac{1}{B}\sum_{\tau\in\mathcal{B}_{j,1}^i} g(\tau|\theta_j^i, \theta_{j-1}^i, P_{j-1}^i)\bigg|\mathcal{F}_j^i\right] = \left(\boldsymbol{\nabla}_\theta \mathbf{J}^H(\theta_{j-1}^i)\right)^\top \boldsymbol{\nabla}_{\mathbf{J}} f(P_{j-1}^i)$$

where $\mathcal{F}_j^i$ represents the randomness before sampling batch $\mathcal{B}_{j,2}^i$. Therefore, we obtain

$$\mathbb{E}\left[\|g_j^i - \left(\boldsymbol{\nabla}_\theta \mathbf{J}^H(\theta_j^i)\right)^\top \boldsymbol{\nabla}_{\mathbf{J}} f(P_j^i)\|^2\right] - \mathbb{E}\left[\|g_{j-1}^i - \left(\boldsymbol{\nabla}_\theta \mathbf{J}^H(\theta_{j-1}^i)\right)^\top \boldsymbol{\nabla}_{\mathbf{J}} f(P_{j-1}^i)\|^2\right]$$
$$\leq \frac{1}{B}\mathbb{E}\left[\|g(\tau|\theta_j^i, P_j^i) - g(\tau|\theta_j^i, \theta_{j-1}^i, P_{j-1}^i)\|^2\right]$$
$$\leq \frac{3}{B}\mathbb{E}\left[\|g(\tau|\theta_j^i, P_j^i) - g(\tau|\theta_{j-1}^i, P_j^i)\|^2\right] + \frac{3}{B}\mathbb{E}\left[\|g(\tau|\theta_{j-1}^i, P_j^i) - g(\tau|\theta_{j-1}^i, P_{j-1}^i)\|^2\right]$$
$$+ \frac{3}{B}\mathbb{E}\left[\|g(\tau|\theta_{j-1}^i, P_{j-1}^i) - g(\tau|\theta_j^i, \theta_{j-1}^i, P_{j-1}^i)\|^2\right].$$

The first term can be upper-bounded as follows:

$$\mathbb{E}\left[\|g(\tau|\theta_j^i, P_j^i) - g(\tau|\theta_{j-1}^i, P_j^i)\|^2\right] \leq \frac{M^2 C^2 S^2}{(1-\gamma)^4}\mathbb{E}\left[\|\theta_j^i - \theta_{j-1}^i\|^2\right]. \quad \text{(by lemma B.1)}$$

Moreover, the second term is upper-bounded as follows:

$$\mathbb{E}\left[\|g(\tau|\theta_{j-1}^i, P_j^i) - g(\tau|\theta_{j-1}^i, P_{j-1}^i)\|^2\right]$$
$$\leq \frac{M^2 G^2 L_f^2}{(1-\gamma)^4}\mathbb{E}\left[\|J_j^i - J_{j-1}^i\|^2\right] \quad \text{(by Lemma B.1 and non-expansiveness of \textbf{Proj})}$$
$$\leq \frac{M^2 G^2 L_f^2}{(1-\gamma)^4 B}\mathbb{E}\left[\|\mathbf{J}(\tau|\theta_{j-1}^i) - \mathbf{J}(\tau|\theta_{j-1}^i, \theta_{j-2}^i)\|^2\right]$$
$$\leq \frac{M^2 G^2 L_f^2}{(1-\gamma)^4 B} \cdot \frac{2MH(2G^2 + S)(\exp\{2GH\delta\} + 1)}{(1-\gamma)^3}\mathbb{E}\left[\|\theta_{j-1}^i - \theta_{j-2}^i\|^2\right]$$
$$= \frac{2M^3 G^2 H L_f^2(2G^2 + S)(\exp\{2GH\delta\} + 1)}{(1-\gamma)^7 B}\mathbb{E}\left[\|\theta_{j-1}^i - \theta_{j-2}^i\|^2\right]$$

Finally, the third term is upper-bounded as follows:

$$\mathbb{E}\left[\|g(\tau|\theta_{j-1}^i, P_{j-1}^i) - g_\omega(\tau|\theta_j^i, \theta_{j-1}^i, P_{j-1}^i)\|^2\right]$$

$$\leq \mathbb{E}\left[\|\sum_{m=1}^{M}(P_{j-1}^i)_m \sum_{t=0}^{H-1}(1-\omega_t(\tau|\theta_j^i, \theta_{j-1}^i))\nabla_\theta \log \pi_{\theta_{j-1}^i}(a_t|s_t) \sum_{h=t}^{H-1}\gamma^h r_m(s_h, a_h)\|^2\right]$$

$$\leq MC^2 \cdot \sum_{m=1}^{M}\mathbb{E}\left[\|\sum_{t=0}^{H-1}(1-\omega_t(\tau|\theta_j^i, \theta_{j-1}^i))\nabla_\theta \log \pi_{\theta_{j-1}^i}(a_t|s_t) \sum_{h=t}^{H-1}\gamma^h r_m(s_h, a_h)\|^2\right]$$

$$\leq MC^2 \cdot \sum_{m=1}^{M}\mathbb{E}\left[\|\sum_{h=0}^{H-1}\gamma^h r_m(s_h, a_h)\sum_{t=0}^{h}(1-\omega_t(\tau|\theta_j^i, \theta_{j-1}^i))\nabla_\theta \log \pi_{\theta_{j-1}^i}(a_t|s_t)\|^2\right]$$

$$\leq MC^2 H \cdot \sum_{m=1}^{M}\mathbb{E}\left[\sum_{h=0}^{H-1}\gamma^h(h+1)\sum_{t=0}^{h}|1-\omega_t(\tau|\theta_j^i, \theta_{j-1}^i)|^2\|\nabla_\theta \log \pi_{\theta_{j-1}^i}(a_t|s_t)\|^2\right]$$

$$\leq MC^2 G^2 H \cdot \sum_{m=1}^{M}\sum_{h=0}^{H-1}\gamma^h(h+1)\sum_{t=0}^{h}\mathbb{V}(\omega_t(\tau|\theta_j^i, \theta_{j-1}^i))$$

$$\leq M^2 C^2 G^2 H \cdot \sum_{h=0}^{H-1}\gamma^h(h+1)\sum_{t=0}^{h}\mathbb{V}(\omega_t(\tau|\theta_j^i, \theta_{j-1}^i))$$

$$\leq M^2 C^2 G^2 H \cdot \mathbb{E}\left[\|\theta_{j-1}^i - \theta_j^i\|^2\right] \cdot \sum_{h=0}^{H-1}\gamma^h h(h+1)^3(2G^2+S)(\exp\{2GH\delta\}+1)$$

$$\leq \frac{24M^2 C^2 G^2 H(2G^2+S)(\exp\{2GH\delta\}+1)}{(1-\gamma)^5} \cdot \mathbb{E}\left[\|\theta_{j-1}^i - \theta_j^i\|^2\right]$$

$\square$

**Lemma C.5**

$$\mathbb{E}\left[\|g_0^i - \nabla_\theta \mathbf{J}^H(\theta_0^i)^\top \nabla_\mathbf{J} f(P_0^i)\|^2\right] \leq \frac{6M^3 G^2 L_f^2}{N(1-\gamma)^6} + \frac{3M^2 C^2 G^2}{N(1-\gamma)^4}$$

*Proof.* Let us define $J_*^i = \mathbf{J}^H(\theta_0^i)$. Then

$$\mathbb{E}\left[\|g_0^i - \left(\nabla_\theta \mathbf{J}^H(\theta_0^i)\right)^\top \nabla_\mathbf{J} f(P_0^i)\|^2\right]$$

$$\leq 3\mathbb{E}\left[\|\frac{1}{N}\sum_{\tau \in \mathcal{N}_2^i}\left(g(\tau|\theta_0^i, P_0^i) - g(\tau|\theta_0^i, J_*^i)\right)\|^2\right] + 3\mathbb{E}\left[\|\frac{1}{N}\sum_{\tau \in \mathcal{N}_2^i}g(\tau|\theta_0^i, J_*^i) - \nabla_\theta f(\mathbf{J}^H(\theta_0^i))\|^2\right]$$

$$+ 3\mathbb{E}\left[\|\nabla_\theta f(\mathbf{J}^H(\theta_0^i)) - \left(\nabla_\theta \mathbf{J}^H(\theta_0^i)\right)^\top \nabla_\mathbf{J} f(P_0^i)\|^2\right].$$

For the first term, we have

$$\mathbb{E}\left[\|\frac{1}{N}\sum_{\tau \in \mathcal{N}_2^i}\left(g(\tau|\theta_0^i, P_0^i) - g(\tau|\theta_0^i, J_*^i)\right)\|^2\right]$$

$$\leq \frac{1}{N}\sum_{\tau \in \mathcal{N}_2^i}\mathbb{E}\left[\|g(\tau|\theta_0^i, P_0^i) - g(\tau|\theta_0^i, J_*^i)\|^2\right]$$

$$\leq \frac{M^2 G^2 L_f^2}{(1-\gamma)^4}\mathbb{E}\left[\|J_0^i - J_*^i\|^2\right] \qquad \text{(by lemma B.1 and non-expansiveness of \textbf{Proj})}$$

$$\leq \frac{M^3 G^2 L_f^2}{N(1-\gamma)^6} \qquad \text{(by lemma C.3)}$$

For the second term, we have

$$\mathbb{E}\left[\|\frac{1}{N}\sum_{\tau\in\mathcal{N}_2^i}g(\tau|\theta_0^i, J_*^i) - \boldsymbol{\nabla}_\theta f(\mathbf{J}^H(\theta_0^i))\|^2\right]$$

$$=\frac{1}{N}\mathbb{E}\left[\|g(\tau|\theta_0^i, J_*^i) - \boldsymbol{\nabla}_\theta f(\mathbf{J}^H(\theta_0^i))\|^2\right] \qquad \text{(by Lemma B.2)}$$

$$\leq \frac{1}{N}\mathbb{E}\left[\|g(\tau|\theta_0^i, J_*^i)\|^2\right]$$

$$=\frac{1}{N}\mathbb{E}\left[\|\sum_{m=1}^{M}\frac{\partial f}{\partial J_m}(J_*^i)\sum_{t=0}^{H-1}\boldsymbol{\nabla}_\theta\log\pi_{\theta_0^i}(a_t|s_t)\sum_{h=t}^{H-1}\gamma^h r_m(s_h, a_h)\|^2\right]$$

$$\leq \frac{M^2 C^2 G^2}{N(1-\gamma)^4}.$$

Finally, for the third term, we have

$$\mathbb{E}\left[\|\boldsymbol{\nabla}_\theta f(\mathbf{J}^H(\theta_0^i)) - \left(\boldsymbol{\nabla}_\theta \mathbf{J}^H(\theta_0^i)\right)^\top \boldsymbol{\nabla}_\mathbf{J} f(P_0^i)\|^2\right]$$

$$=\mathbb{E}\left[\|\left(\boldsymbol{\nabla}_\theta \mathbf{J}^H(\theta_0^i)\right)^\top \left(\boldsymbol{\nabla}_\mathbf{J} f(J_*^i) - \boldsymbol{\nabla}_\mathbf{J} f(P_0^i)\right)\|^2\right]$$

$$\leq \mathbb{E}\left[\|\boldsymbol{\nabla}_\theta \mathbf{J}^H(\theta_0^i)\|^2 \|\boldsymbol{\nabla}_\mathbf{J} f(J_*^i) - \boldsymbol{\nabla}_\mathbf{J} f(P_0^i)\|^2\right]$$

$$\leq \frac{MG^2}{(1-\gamma)^4}\mathbb{E}\left[\|\boldsymbol{\nabla}_\mathbf{J} f(J_*^i) - \boldsymbol{\nabla}_\mathbf{J} f(P_0^i)\|^2\right] \qquad \text{(by Lemma B.1)}$$

$$\leq \frac{M^3 G^2 L_f^2}{(1-\gamma)^6 N}. \qquad \text{(by Corollary 1)}$$

□

### C.2.3 End of Proof

*Proof of Lemma 5.*

$$\mathbb{E}\left[\|g_j^i - \boldsymbol{\nabla}_\theta f(\mathbf{J}(\theta_j^i))\|^2\right]$$

$$\leq 3\mathbb{E}\left[\|g_j^i - \boldsymbol{\nabla}_\theta \mathbf{J}^H(\theta_j^i)^\top \boldsymbol{\nabla}_\mathbf{J} f(P_j^i)\|^2\right] + 3\mathbb{E}\left[\|\boldsymbol{\nabla}_\theta \mathbf{J}^H(\theta_j^i)^\top \boldsymbol{\nabla}_\mathbf{J} f(P_j^i) - \boldsymbol{\nabla}_\theta f(\mathbf{J}^H(\theta_j^i))\|^2\right]$$

$$+ 3\mathbb{E}\left[\boldsymbol{\nabla}_\theta f(\mathbf{J}^H(\theta_j^i)) - \boldsymbol{\nabla}_\theta f(\mathbf{J}(\theta_j^i))\|^2\right]$$

The first term is upper-bounded using Lemmas C.4 and C.5.
The second term is upper-bounded as follows:

$$\mathbb{E}\left[\|\boldsymbol{\nabla}_\theta \mathbf{J}^H(\theta_j^i)^\top \left(\boldsymbol{\nabla}_\mathbf{J} f(P_j^i) - \boldsymbol{\nabla}_\mathbf{J} f(\mathbf{J}^H(\theta_j^i))\right)\|^2\right]$$

$$\leq \frac{MG^2}{(1-\gamma)^4}\cdot\mathbb{E}\left[\|\boldsymbol{\nabla}_\mathbf{J} f(P_j^i) - \boldsymbol{\nabla}_\mathbf{J} f(\mathbf{J}^H(\theta_j^i))\|^2\right] \qquad \text{(by Corollary 1)}$$

$$\leq \frac{M^2 G^2 L_f^2}{(1-\gamma)^4}\cdot\mathbb{E}\left[\|J_j^i - \mathbf{J}^H(\theta_j^i)\|^2\right] \qquad \text{(by Assumption 3 and non-expansiveness of \textbf{Proj})}$$

$$\leq \frac{M^2 G^2 L_f^2}{(1-\gamma)^4}\cdot\left(\frac{M}{(1-\gamma)^2 N} + \frac{2MH(2G^2+S)(\exp\{2GH\delta\}+1)}{(1-\gamma)^3 B}\cdot\sum_{j'=1}^{j}\mathbb{E}\left[\|\theta_{j-1}^i - \theta_j^i\|^2\right]\right)$$

(by Lemma C.3)

The last term is upper bounded directly by lemma 2.
Combining the inequalities we obtain

$$\mathbb{E}\left[\|g_j^i - \boldsymbol{\nabla}_\theta f(\mathbf{J}(\theta_j^i))\|^2\right] \leq \frac{C_1}{N} + C_2\gamma^{2H} + \frac{C_3}{B}\sum_{j'=1}^{j}\mathbb{E}\left[\|\theta_{j-1}^i - \theta_j^i\|^2\right]$$

where
$$C_1 = \frac{21M^3G^2L_f^2}{(1-\gamma^6)} + \frac{9M^2C^2G^2}{(1-\gamma)^4},$$
$$C_2 = \frac{3M^2G^2}{(1-\gamma)^4}\left[\sqrt{M}L_f\frac{1-\gamma^H - H\gamma^H(1-\gamma)}{1-\gamma} + C[1+H(1-\gamma)]\right]^2,$$
$$C_3 = \frac{9M^2C^2S^2}{(1-\gamma)^4} + \frac{18M^2G^2H(2G^2+S)(\exp\{2GH\delta\}+1)}{(1-\gamma)^5}\left(12C^2 + \frac{4ML_f^2}{3(1-\gamma)^2}\right).$$
□

## D Ascent Lemmas (for Stationary Convergence)

**Lemma D.1**
Let Assumptions 1, 2, 3 hold. The following ascent inequality holds for Algorithm 1
$$f(\mathbf{J}(\theta_{i+1})) \geq f(\mathbf{J}(\theta_i)) + \frac{\eta}{4}\|\nabla_\theta f(\mathbf{J}(\theta_i))\|^2 +$$
$$+ \left(\frac{1}{2\eta} - L_\theta\right)\|\theta_{i+1}-\theta_i\|^2 - \left(\frac{\eta}{2} + \frac{1}{2L_\theta}\right)\|\nabla_\theta f(\mathbf{J}(\theta_i)) - g_i\|^2.$$

*Proof.* Let us denote
$$\hat{\theta}_{i+1} = \theta_i + \eta \cdot \nabla_\theta f(\mathbf{J}(\theta_i)).$$
Now, we have
$$\|\nabla_\theta f(\mathbf{J}(\theta_i))\|^2 = \eta^{-2}\|\hat{\theta}_{i+1} - \theta_i\|^2$$
$$\leq 2\eta^{-2}\|\theta_{i+1} - \theta_i\|^2 + 2\eta^{-2}\|\hat{\theta}_{i+1} - \theta_{i+1}\|$$
$$\leq 2\eta^{-2}\|\theta_{i+1} - \theta_i\|^2 + 2\|g_i - \nabla_\theta f(\mathbf{J}(\theta_i))\|^2. \qquad (2)$$

Therefore, by $L_\theta$-smoothness of the objective function, we obtain

$f(\mathbf{J}(\theta_i))$
$$\geq f(\mathbf{J}(\theta_i)) + \langle \nabla_\theta f(\mathbf{J}(\theta_i)), \theta_{i+1} - \theta_i\rangle - \frac{L_\theta}{2}\|\theta_{i+1}-\theta_i\|^2$$
$$= f(\mathbf{J}(\theta_i)) + \langle g_i, \theta_{i+1} - \theta_i\rangle - \frac{1}{2\eta}\|\theta_{i+1}-\theta_i\|^2 + \langle \nabla_\theta f(\mathbf{J}(\theta_i)) - g_i, \theta_{i+1}-\theta_i\rangle$$
$$+ \left(\frac{1}{2\eta} - \frac{L_\theta}{2}\right)\|\theta_{i+1}-\theta_i\|^2$$
$$\stackrel{(a)}{=} f(\mathbf{J}(\theta_i)) + \langle \nabla_\theta f(\mathbf{J}(\theta_i)) - g_i, \theta_{i+1}-\theta_i\rangle + \left(\frac{1}{\eta} - \frac{L_\theta}{2}\right)\|\theta_{i+1}-\theta_i\|^2$$
$$\stackrel{(b)}{\geq} f(\mathbf{J}(\theta_i)) + \frac{\eta}{4}\|\mathcal{G}_\eta(\theta_i)\|^2 + \left(\frac{1}{2\eta} - L_\theta\right)\|\theta_{i+1}-\theta_i\|^2 - \left(\frac{\eta}{2} + \frac{1}{2L_\theta}\right)\|\nabla_\theta f(\mathbf{J}(\theta_i)) - g_i\|^2,$$

where (a) is due to
$$\langle g_i, \theta_{i+1} - \theta_i\rangle = \frac{1}{\eta}\|\theta_{i+1}-\theta_i\|^2$$
by the update rule; whereas (b) is due to
$$\langle \nabla_\theta f(\mathbf{J}(\theta_i)) - g_i, \theta_{i+1}-\theta_i\rangle \geq -\left(\frac{1}{L_\theta}\|\nabla_\theta f(\mathbf{J}(\theta_i)) - g_i\|\right) \cdot (L_\theta\|\theta_{i+1}-\theta_i\|)$$
$$\text{(by Cauchy's inequality)}$$
$$\geq -\frac{1}{2L_\theta}\|\nabla_\theta f(\mathbf{J}(\theta_i)) - g_i\|^2 - \frac{L_\theta}{2}\|\theta_{i+1}-\theta_i\|$$
$$\text{(since } 2ab \leq a^2 + b^2\text{)}$$

and adding
$$0 \geq \frac{\eta}{4}\|\nabla_\theta f(\mathbf{J}(\theta_i))\|^2 - \frac{1}{2\eta}\|\theta_{i+1}-\theta_i\|^2 - \frac{\eta}{2}\|\nabla_\theta f(\mathbf{J}(\theta_i)) - g_i\|^2$$
by inequality 2. □

**Lemma D.2**
Under the same setting as Lemma D.1, the following ascent inequality holds for Algorithm 2

$$f(\mathbf{J}(\theta_{j+1}^i)) \geq f(\mathbf{J}(\theta_j^i)) + \frac{\eta}{4}\|\mathcal{G}_\eta(\theta_j^i)\|^2 +$$
$$+ \left(\frac{1}{2\eta} - L_\theta\right)\|\theta_{j+1}^i - \theta_j^i\|^2 - \left(\frac{\eta}{2} + \frac{1}{2L_\theta}\right)\|\nabla_\theta f(\mathbf{J}(\theta_j^i)) - g_j^i\|^2.$$

where $\mathcal{G}_\eta(\theta) := \frac{1}{\eta}\left(\mathbf{Proj}_{B(\theta_j^i,\delta)}(\theta + \eta \cdot \nabla_\theta f(\mathbf{J}(\theta))) - \theta\right)$.

*Proof.* For algorithm 2, the truncated gradient update iterates can also be written as $\theta_{j+1}^i = \mathbf{Proj}_{B(\theta_j^i,\delta)}(\theta_j^i + \eta \cdot g_j^i)$. Moreover, let us denote

$$\hat{\theta}_{j+1}^i = \mathbf{Proj}_{B(\theta_j^i,\delta)}\left(\theta_j^i + \eta \cdot \nabla_\theta f(\mathbf{J}(\theta_j^i))\right).$$

Now, we have

$$\begin{aligned}
\|\mathcal{G}_\eta(\theta_j^i)\|^2 &= \eta^{-2}\|\hat{\theta}_{j+1}^i - \theta_j^i\|^2 \\
&\leq 2\eta^{-2}\|\theta_{j+1}^i - \theta_j^i\|^2 + 2\eta^{-2}\|\hat{\theta}_{j+1}^i - \theta_{j+1}^i\| \\
&= 2\eta^{-2}\|\theta_{j+1}^i - \theta_j^i\|^2 + 2\eta^{-2}\|\mathbf{Proj}_{B(\theta_j^i,\delta)}(\theta_j^i + \eta g_j^i) - \mathbf{Proj}_{B(\theta_j^i,\delta)}\left(\theta_i^j + \eta\nabla_\theta f(\mathbf{J}(\theta_j^i))\right)\|^2 \\
&\leq 2\eta^{-2}\|\theta_{j+1}^i - \theta_j^i\|^2 + 2\|g_j^i - \nabla_\theta f(\mathbf{J}(\theta_j^i))\|^2. \qquad (3)
\end{aligned}$$

Therefore, by $L_\theta$-smoothness of the objective function, we obtain

$$\begin{aligned}
&f(\mathbf{J}(\theta_{j+1}^i)) \\
&\geq f(\mathbf{J}(\theta_j^i)) + \langle\nabla_\theta f(\mathbf{J}(\theta_j^i)), \theta_{j+1}^i - \theta_j^i\rangle - \frac{L_\theta}{2}\|\theta_{j+1}^i - \theta_j^i\|^2 \\
&= f(\mathbf{J}(\theta_j^i)) + \langle g_j^i, \theta_{j+1}^i - \theta_j^i\rangle - \frac{1}{2\eta}\|\theta_{j+1}^i - \theta_j^i\|^2 + \langle\nabla_\theta f(\mathbf{J}(\theta_j^i)) - g_j^i, \theta_{j+1}^i - \theta_j^i\rangle \\
&\quad + \left(\frac{1}{2\eta} - \frac{L_\theta}{2}\right)\|\theta_{j+1}^i - \theta_j^i\|^2 \\
&\stackrel{(a)}{\geq} f(\mathbf{J}(\theta_j^i)) + \langle\nabla_\theta f(\mathbf{J}(\theta_j^i)) - g_j^i, \theta_{j+1}^i - \theta_j^i\rangle + \left(\frac{1}{\eta} - \frac{L_\theta}{2}\right)\|\theta_{j+1}^i - \theta_j^i\|^2 \\
&\stackrel{(b)}{\geq} f(\mathbf{J}(\theta_j^i)) + \frac{\eta}{4}\|\mathcal{G}_\eta(\theta_j^i)\|^2 + \left(\frac{1}{2\eta} - L_\theta\right)\|\theta_{j+1}^i - \theta_j^i\|^2 - \left(\frac{\eta}{2} + \frac{1}{2L_\theta}\right)\|\nabla_\theta f(\mathbf{J}(\theta_j^i)) - g_j^i\|^2,
\end{aligned}$$

where (a) is due to

$$\langle g_j^i, \theta_{j+1}^i - \theta_j^i\rangle \geq \frac{1}{\eta}\|\theta_{j+1}^i - \theta_j^i\|^2$$

by the update rule; whereas (b) is due to

$$\begin{aligned}
\langle\nabla_\theta f(\mathbf{J}(\theta_j^i)) - g_j^i, \theta_{j+1}^i - \theta_j^i\rangle &\geq -\left(\frac{1}{L_\theta}\|\nabla_\theta f(\mathbf{J}(\theta_j^i)) - g_j^i\|\right) \cdot \left(L_\theta\|\theta_{j+1}^i - \theta_j^i\|\right) \\
&\quad \text{(by Cauchy's inequality)} \\
&\geq -\frac{1}{2L_\theta}\|\nabla_\theta f(\mathbf{J}(\theta_j^i)) - g_j^i\|^2 - \frac{L_\theta}{2}\|\theta_{j+1}^i - \theta_j^i\| \\
&\quad \text{(since } 2ab \leq a^2 + b^2\text{)}
\end{aligned}$$

and adding

$$0 \geq \frac{\eta}{4}\|\mathcal{G}_\eta(\theta_j^i)\|^2 - \frac{1}{2\eta}\|\theta_{j+1}^i - \theta_j^i\|^2 - \frac{\eta}{2}\|\nabla_\theta f(\mathbf{J}(\theta_j^i)) - g_j^i\|^2$$

by inequality 3. □

# E Stationary Convergence

## E.1 Proof of Theorem 1

*Proof.* From Lemma D.1 and lemma 4, we obtain

$$\frac{\eta}{4}\sum_{i=0}^{T-1}\mathbb{E}\left[\|\boldsymbol{\nabla}_\theta f(\mathbf{J}(\theta_i))\|^2\right]$$

$$\leq \mathbb{E}\left[f(\mathbf{J}(\theta_T))\right] - \mathbb{E}\left[f(\mathbf{J}(\theta_0))\right] - \left(\frac{1}{2\eta} - L_\theta\right)\sum_{i=0}^{T-1}\mathbb{E}\left[\|\theta_{i+1} - \theta_i\|^2\right]$$

$$+ \left(\frac{\eta}{2} + \frac{1}{2L_\theta}\right)\sum_{i=0}^{T-1}\mathbb{E}\left[\|\nabla_\theta F(\mathbf{J}(\theta_i)) - g_i\|^2\right]$$

$$\leq \mathbb{E}\left[f(\mathbf{J}(\theta^*))\right] - \mathbb{E}\left[f(\mathbf{J}(\theta_0))\right] + T\left(\frac{\eta}{2} + \frac{1}{2L_\theta}\right)\left(\frac{C_1}{N} + \gamma^{2H}C_2\right)$$

$$- \left(\frac{1}{2\eta} - L_\theta\right)\sum_{i=0}^{T-1}\mathbb{E}\left[\|\theta_{i+1} - \theta_i\|^2\right].$$

Note that

$$\frac{1}{2\eta} - L_\theta = 0$$

since $\eta = \frac{1}{2L_\theta}$, therefore we have

$$\frac{\eta}{4}\sum_{i=0}^{T-1}\mathbb{E}\left[\|\boldsymbol{\nabla}_\theta f(\theta_i)\|^2\right] \leq \mathbb{E}\left[f(\mathbf{J}(\theta^*))\right] - \mathbb{E}\left[f(\mathbf{J}(\theta_0))\right] + T\left(\frac{\eta}{2} + \frac{1}{2L_\theta}\right)\left(\frac{C_1}{N} + \gamma^{2H}C_2\right).$$

Since $T = M\varepsilon^{-2}$, $N = M^3\varepsilon^{-2}$, $H = \frac{6\log(M\varepsilon^{-1})}{1-\gamma}$, the constants introduced in Lemma 5 satisfy $C_1, C_2, C_3 \in \mathcal{O}(M^3)$, omitting the lipshitz constants and the dependence on $(1-\gamma)^{-1}$. Moreover, we have $\eta \in \Theta(M^{-1})$ and $L_\theta \in \mathcal{O}(M)$. Let $\theta_{out}$ be selected uniformly at random from $\{\theta_i\}_{i=0,\dots,T-1}$, then we obtain

$$\mathbb{E}\left[\|\boldsymbol{\nabla}_\theta f(\mathbf{J}(\theta_{out}))\|^2\right]$$

$$= \frac{1}{T}\sum_{i=0}^{T-1}\mathbb{E}\left[\|\boldsymbol{\nabla}_\theta f(\mathbf{J}(\theta_i))\|^2\right]$$

$$\leq \frac{4\mathbb{E}\left[f(\mathbf{J}(\theta^*)) - f(\mathbf{J}(\theta_0))\right]}{T \cdot \eta} + 6\left(\frac{C_1}{N} + \gamma^{2H}C_2\right)$$

$$\leq \left(4(M\eta)^{-1}\mathbb{E}\left[f(\mathbf{J}(\theta^*)) - f(\mathbf{J}(\theta_0))\right] + 6\left(\frac{C_1}{M^3} + \gamma^{2H}\varepsilon^{-2}C_2\right)\right) \cdot \varepsilon^2$$

$$= (\text{constant w.r.t. } \varepsilon^{-1} \text{ and } M \text{ up to log factors}) \cdot \varepsilon^2$$

$$= \tilde{\mathcal{O}}(\varepsilon^2).$$

Then, by Jensen's inequality, we have $\mathbb{E}\left[\|\boldsymbol{\nabla}_\theta f(\mathbf{J}(\theta_{out}))\|\right] \leq \tilde{\mathcal{O}}(\varepsilon)$.

□

### E.2 Proof of Theorem 2

*Proof.* From the Lemmas D.2 and 5, we obtain

$$\frac{\eta}{4} \sum_{j=0}^{m-1} \mathbb{E}\left[\|\mathcal{G}_\eta(\theta_j^i)\|^2\right]$$

$$\leq \mathbb{E}\left[f(\mathbf{J}(\theta_m^i))\right] - \mathbb{E}\left[f(\mathbf{J}(\theta_0^i))\right] - \left(\frac{1}{2\eta} - L_\theta\right) \sum_{j=0}^{m-1} \mathbb{E}\left[\|\theta_{j+1}^i - \theta_j^i\|^2\right]$$

$$+ \left(\frac{\eta}{2} + \frac{1}{2L_\theta}\right) \sum_{j=0}^{m-1} \mathbb{E}\left[\|\nabla_\theta F(\mathbf{J}(\theta_j^i)) - g_j^i\|^2\right]$$

$$\leq \mathbb{E}\left[f(\mathbf{J}(\theta_m^i))\right] - \mathbb{E}\left[f(\mathbf{J}(\theta_0^i))\right] + m\left(\frac{\eta}{2} + \frac{1}{2L_\theta}\right)\left(\frac{C_1}{N} + \gamma^{2H}C_2\right)$$

$$- \left(\left(\frac{1}{2\eta} - L_\theta\right) - \left(\frac{\eta}{2} + \frac{1}{2L_\theta}\right)\frac{m}{B}C_3\right)\sum_{j=0}^{m-1} \mathbb{E}\left[\|\theta_{j+1}^i - \theta_j^i\|^2\right].$$

Note that

$$\left(\frac{1}{2\eta} - L_\theta\right) - \left(\frac{\eta}{2} + \frac{1}{2L_\theta}\right)\frac{m}{B}C_3 = \frac{C_3 L_\theta^2 M + 2C_3^2}{4C_3 L_\theta M + 4L_\theta^3 M^2} \geq 0$$

since $B = M^2 \varepsilon^{-1}, m = M\varepsilon^{-1}$ and $\eta = \frac{1}{1+C_3/(ML_\theta^2)} \cdot \frac{1}{2L_\theta}$, therefore we have

$$\frac{\eta}{4} \sum_{j=0}^{m-1} \mathbb{E}\left[\|\mathcal{G}_\eta(\theta_j^i)\|^2\right] \leq \mathbb{E}\left[f(\mathbf{J}(\theta_m^i))\right] - \mathbb{E}\left[f(\mathbf{J}(\theta_0^i))\right] + m\left(\frac{\eta}{2} + \frac{1}{2L_\theta}\right)\left(\frac{C_1}{N} + \gamma^{2H}C_2\right)$$

Summing the above inequality over all $T$ epochs and dividing both sides by $\frac{\eta}{4}Tm$ yields

$$\frac{1}{Tm} \sum_{j=0}^{m-1} \sum_{i=1}^{T} \mathbb{E}\left[\|\mathcal{G}_\eta(\theta_j^i)\|^2\right] \leq \frac{4\left(f(\mathbf{J}(\theta^*)) - f(\mathbf{J}(\tilde{\theta}_0))\right)}{Tm \cdot \eta} + (2 + \frac{2}{L_\theta \eta})(\frac{C_1}{N} + \gamma^{2H}C_2)$$

Since $T = \varepsilon^{-1}$, $N = M^3\varepsilon^{-2}$, $H = \frac{6\log(M\varepsilon^{-1})}{1-\gamma}$, $\delta = \frac{1}{2GH}$, the constants introduced in Lemma 5 satisfy $C_1, C_2, C_3 \in \mathcal{O}(M^3)$, omitting the lipshitz constants and the dependence on $(1-\gamma)^{-1}$. Moreover, we have $\eta \in \Theta(M^{-1})$ and $L_\theta \in \mathcal{O}(M)$. Let $\theta_{out}$ be selected uniformly at random from $\{\theta_j^i\}_{j=0,\ldots,m-1}^{i=1,\ldots,T}$, then we obtain

$$\mathbb{E}\left[\|\mathcal{G}_\eta(\theta_{out})\|^2\right]$$

$$= \frac{1}{Tm} \sum_{j=0}^{m-1} \sum_{i=1}^{T} \mathbb{E}\left[\|\mathcal{G}_\eta(\theta_j^i)\|^2\right]$$

$$\leq \frac{4\mathbb{E}\left[f(\mathbf{J}(\theta^*)) - f(\mathbf{J}(\tilde{\theta}_0))\right]}{Tm \cdot \eta} + \left(2 + \frac{2}{L_\theta \eta}\right)\left(\frac{C_1}{N} + \gamma^{2H}C_2\right)$$

$$\leq \left(4(M\eta)^{-1}\mathbb{E}\left[f(\mathbf{J}(\theta^*)) - f(\mathbf{J}(\tilde{\theta}_0))\right] + \left(6 + 4\frac{C_3}{ML_\theta^2}\right)\left(\frac{C_1}{M^3} + \gamma^{2H}\varepsilon^{-2}C_2\right)\right) \cdot \varepsilon^2$$

$$= (\text{constant w.r.t. } \varepsilon^{-1} \text{ and } M \text{ up to log factors}) \cdot \varepsilon^2$$

$$= \tilde{\mathcal{O}}(\varepsilon^2).$$

Then, by Jensen's inequality, we have $\mathbb{E}\left[\|\nabla_\theta f(\mathbf{J}(\theta_{out}))\|\right] \leq \tilde{\mathcal{O}}(\varepsilon)$. Hence, we achieve $\mathbb{E}\left[\|\nabla_\theta f(\mathbf{J}(\theta_{out}))\|^2\right] \leq \tilde{\mathcal{O}}(\varepsilon^2)$ by Lemma B.3, which does not introduce any extra dependence on $M$ since $M\eta = \mathcal{O}(1)$.

□

# F  Ascent Lemmas (Global Convergence)

**Lemma F.1**
Let Assumptions 1, 2, 3, 4 and 5 be satisfied, the following result holds for Algorithm 1

$$f(\mathbf{J}(\theta^*)) - f(\mathbf{J}(\theta_{i+1}))$$
$$\leq (1-\varepsilon)(f(\mathbf{J}(\theta^*)) - f(\mathbf{J}(\theta_i))) + \left(L_\theta + \frac{1}{2\eta}\right)\frac{4Ml_\theta^2}{(1-\gamma)^2}\varepsilon^2 - \left(\frac{1}{2\eta} - L_\theta\right)\|\theta_{i+1} - \theta_i\|^2$$
$$+ \frac{1}{L_\theta}\|g_i - \boldsymbol{\nabla}_\theta f(\mathbf{J}(\theta_i))\|^2$$

for all $\varepsilon \leq \bar{\varepsilon}$.

*Proof.* Let $\varepsilon \leq \bar{\varepsilon}$. By $L_\theta$ smoothness of $f(\mathbf{J}(\cdot))$, we have

$$|f(\mathbf{J}(\theta)) - f(\mathbf{J}(\theta_i)) - \langle g_i, \theta - \theta_i \rangle| \leq L_\theta \|\theta - \theta_i\|^2 + \frac{1}{2L_\theta}\|g_i - \boldsymbol{\nabla}_\theta f(\mathbf{J}(\theta_i))\|^2.$$

Moreover, note that the update step of Algorithm 1 satisfies the following:

$$\theta_{i+1} = \arg\max_{\theta \in \Theta} f(\mathbf{J}(\theta_i)) + \langle g_i, \theta - \theta_i \rangle - \frac{1}{2\eta}\|\theta - \theta_i\|^2.$$

From the above two inequalities, it follows that

$$f(\mathbf{J}(\theta_i)) \geq \max_{\theta \in \Theta}\left\{f(\mathbf{J}(\theta)) - \left(L_\theta + \frac{1}{2\eta}\right)\right\} + \left(\frac{1}{2\eta} - L_\theta\right)\|\theta_{i+1} - \theta_i\|^2 \qquad (4)$$
$$- \frac{1}{L_\theta}\|g_i - \boldsymbol{\nabla}_\theta f(\mathbf{J}(\theta_i))\|^2.$$

Consider $\theta_\varepsilon := (\mathbf{J}|_{\mathcal{U}_{\theta_i}})^{-1}((1-\varepsilon)\mathbf{J}(\theta_i) + \varepsilon \mathbf{J}(\theta^*))$, substituting in inequality 4 yielding

$$f(\mathbf{J}(\theta_{i+1})) \geq f((1-\varepsilon)\mathbf{J}(\theta_i) + \varepsilon \mathbf{J}(\theta^*)) - \left(L_\theta + \frac{1}{2\eta}\right)\|\theta_\varepsilon - \theta_i\|^2$$
$$+ \left(\frac{1}{2\eta} - L_\theta\right)\|\theta_{i+1} - \theta_i\|^2 - \frac{1}{L_\theta}\|g_i - \boldsymbol{\nabla}_\theta f(\mathbf{J}(\theta_i))\|^2.$$

Applying concavity of $f$ on the first term of the RHS and noting that

$$\|\theta_\varepsilon - \theta_i\|^2 \leq \varepsilon^2 l_\theta^2 \|\mathbf{J}(\theta_i) - \mathbf{J}(\theta^*)\|^2 \qquad \text{(by Assumption 5)}$$
$$\leq 2\varepsilon^2 l_\theta^2 (\|\mathbf{J}(\theta_i)\|^2 + \|\mathbf{J}(\theta^*)\|^2)$$
$$\leq \frac{4M\varepsilon^2 l_\theta^2}{(1-\gamma)^2},$$

we obtain the desired inequality. $\square$

**Lemma F.2**
Under the same setting as Lemma D.1, the following result holds for Algorithm 2

$$f(\mathbf{J}(\theta^*)) - f(\mathbf{J}(\theta_{j+1}^i))$$
$$\leq (1-\varepsilon)(f(\mathbf{J}(\theta^*)) - f(\mathbf{J}(\theta_j^i))) + \left(L_\theta + \frac{1}{2\eta}\right)\frac{4Ml_\theta^2}{(1-\gamma)^2}\varepsilon^2 - \left(\frac{1}{2\eta} - L_\theta\right)\|\theta_{j+1}^i - \theta_j^i\|^2$$
$$+ \frac{1}{L_\theta}\|g_j^i - \boldsymbol{\nabla}_\theta f(\mathbf{J}(\theta_j^i))\|^2$$

for all $\varepsilon \leq \bar{\varepsilon}$.

*Proof.* Let $\varepsilon \leq \bar{\varepsilon}$. By $L_\theta$ smoothness of $f(\mathbf{J}(\cdot))$, we have

$$|f(\mathbf{J}(\theta)) - f(\mathbf{J}(\theta_j^i)) - \langle g_j^i, \theta - \theta_j^i \rangle| \leq L_\theta \|\theta - \theta_j^i\|^2 + \frac{1}{2L_\theta}\|g_j^i - \boldsymbol{\nabla}_\theta f(\mathbf{J}(\theta_j^i))\|^2.$$

Moreover, note that the update step of Algorithm 2 is equivalent to solving the following trust-region problem:

$$\theta_{j+1}^i = \arg\max_{\theta \in B(\theta_j^i, \delta)} f(\mathbf{J}(\theta_j^i)) + \langle g_j^i, \theta - \theta_j^i \rangle - \frac{1}{2\eta} \|\theta - \theta_j^i\|^2.$$

From the above two inequalities, it follows that

$$f(\mathbf{J}(\theta_j^i)) \geq \max_{\theta \in B(\theta_j^i, \delta)} \left\{ f(\mathbf{J}(\theta)) - \left(L_\theta + \frac{1}{2\eta}\right) \right\} + \left(\frac{1}{2\eta} - L_\theta\right) \|\theta_{j+1}^i - \theta_j^i\|^2 \quad (5)$$
$$- \frac{1}{L_\theta} \|g_j^i - \nabla_\theta f(\mathbf{J}(\theta_j^i))\|^2.$$

Consider $\theta_\varepsilon := (\mathbf{J}|_{\mathcal{U}_{\theta_j^i}})^{-1}((1-\varepsilon)\mathbf{J}(\theta_j^i) + \varepsilon \mathbf{J}(\theta^*))$. Note that $\theta_\varepsilon \subseteq B(\theta_j^i, \theta)$ by Assumption 5, therefore it can be substituted in the inequality 5 yielding

$$f(\mathbf{J}(\theta_{j+1}^i)) \geq f((1-\varepsilon)\mathbf{J}(\theta_j^i) + \varepsilon \mathbf{J}(\theta^*)) - \left(L_\theta + \frac{1}{2\eta}\right) \|\theta_\varepsilon - \theta_j^i\|^2$$
$$+ \left(\frac{1}{2\eta} - L_\theta\right) \|\theta_{j+1}^i - \theta_j^i\|^2 - \frac{1}{L_\theta} \|g_j^i - \nabla_\theta f(\mathbf{J}(\theta_j^i))\|^2.$$

Applying concavity of $f$ on the first term of the RHS and noting that

$$\|\theta_\varepsilon - \theta_j^i\|^2 \leq \varepsilon^2 l_\theta^2 \|\mathbf{J}(\theta_j^i) - \mathbf{J}(\theta^*)\|^2 \quad \text{(by Assumption 5)}$$
$$\leq 2\varepsilon^2 l_\theta^2 (\|\mathbf{J}(\theta_j^i)\|^2 + \|\mathbf{J}(\theta^*)\|^2)$$
$$\leq \frac{4M\varepsilon^2 l_\theta^2}{(1-\gamma)^2},$$

we obtain the desired inequality. □

## G  Global Convergence

### G.1  Proof of Theorem 3

*Proof.* Let $\sigma_i = -\left(\frac{1}{2\eta} - L_\theta\right) \|\theta_{i+1} - \theta_i\|^2 + \frac{1}{L_\theta} \|g_i - \nabla_\theta f(\mathbf{J}(\theta_i))\|^2$. Unrolling the recursive inequality showed in Lemma F.1 and taking the expectation of both sides, we obtain

$$\mathbb{E}\left[f(\mathbf{J}(\theta^*)) - f(\mathbf{J}(\theta_T))\right] \leq (1-\varepsilon)^T \mathbb{E}\left[f(\mathbf{J}(\theta^*)) - f(\mathbf{J}(\theta_0))\right] + \frac{(4L_\theta + \frac{2}{\eta})l_\theta^2}{M(1-\gamma)^2}\varepsilon$$
$$+ \sum_{i=0}^{T-1} (1-\varepsilon)^{T-i-1} \mathbb{E}[\sigma_i]$$

for all $\varepsilon \leq \bar{\varepsilon}$. Moreover, substituting $\varepsilon \to \frac{\varepsilon}{M^2}$, we arrive at

$$\mathbb{E}\left[f(\mathbf{J}(\theta^*)) - f(\mathbf{J}(\theta_T))\right] \leq \left(1 - \frac{\varepsilon}{M^2}\right)^T \mathbb{E}\left[f(\mathbf{J}(\theta^*)) - f(\mathbf{J}(\theta_0))\right]$$
$$+ \frac{(4L_\theta + \frac{2}{\eta})l_\theta^2}{M(1-\gamma)^2}\varepsilon + \sum_{i=0}^{T-1} \left(1 - \frac{\varepsilon}{M^2}\right)^{T-i-1} \mathbb{E}[\sigma_i]$$

for all $\varepsilon \leq \bar{\varepsilon}$.

Next, note that by choosing $T = M^2\varepsilon^{-1}\ln(\varepsilon^{-1}), N = M^4\varepsilon^{-2}\ln(\varepsilon^{-1})$ and $\eta = \frac{1}{2L_\theta + 8C_3/(ML_\theta)}$ and if $\varepsilon \leq \frac{1}{2}$, we compute

$$\sum_{i=0}^{T-1}(1-\frac{\varepsilon}{M^2})^{T-i-1}\mathbb{E}[\sigma_i]$$

$$\leq \frac{1}{L_\theta}\sum_{i=0}^{T-1}\mathbb{E}\left[\|g_i - \nabla_\theta f(\mathbf{J}(\theta_i))\|^2\right]$$

$$- \left(1-\frac{\varepsilon}{M^2}\right)^{M^2\varepsilon^{-1}\ln(\varepsilon^{-1})}\left(\frac{1}{2\eta} - L_\theta\right)\sum_{i=0}^{T-1}\mathbb{E}\left[\|\theta_{i+1} - \theta_i\|^2\right]$$

$$\overset{(i)}{\leq} \frac{1}{L_\theta}\sum_{i=0}^{T-1}\mathbb{E}\left[\|g_i - \nabla_\theta f(\mathbf{J}(\theta_i))\|^2\right] - \left(\frac{1}{8\eta} - \frac{L_\theta}{4}\right)\sum_{j=0}^{T-1}\mathbb{E}\left[\|\theta_{i+1} - \theta_i\|^2\right]$$

$$\overset{(ii)}{\leq} \frac{TC_1}{L_\theta N} + \frac{TC_2\gamma^{2H}}{L_\theta} - \frac{C_3}{ML_\theta}\sum_{j=0}^{T-1}\mathbb{E}\left[\|\theta_{i+1} - \theta_i\|^2\right]$$

$$\leq \frac{C_1}{L_\theta M^2}\varepsilon + \frac{M^2 C_2 \ln(\varepsilon^{-1})}{L_\theta \varepsilon}\gamma^{2H}$$

where step (i) holds since $(1 - \frac{\varepsilon}{M^2})^{M^2\varepsilon^{-1}\ln(\varepsilon^{-1})} \geq \frac{1}{4}$ and step (ii) follows by Lemma 5. Therefore, noting that $(1 - \frac{\varepsilon}{M^2})^{M^2\varepsilon^{-1}\ln(\varepsilon^{-1})} \leq \varepsilon$ for $\varepsilon \in [0, \frac{1}{2}]$, we have

$$\mathbb{E}[f(\mathbf{J}(\theta^*)) - f(\mathbf{J}(\theta_T))] \leq \varepsilon \mathbb{E}[f(\mathbf{J}(\theta^*)) - f(\mathbf{J}(\theta_0))] \qquad (6)$$
$$+ \left(\frac{(4L_\theta + \frac{2}{\eta})l_\theta^2}{M(1-\gamma)^2} + \frac{C_1}{M^2 L_\theta}\right)\varepsilon + \frac{M^2 C_2 \ln(\varepsilon^{-1})}{L_\theta \varepsilon}\gamma^{2H}.$$

Since $H = \frac{\ln(M\varepsilon^{-1})}{1-\gamma}$, we conclude $\mathbb{E}[f(\mathbf{J}(\theta^*)) - f(\mathbf{J}(\theta_T))] \leq \tilde{\mathcal{O}}(\varepsilon)$, as desired. $\square$

### G.2 Proof of Theorem 4

*Proof.* Let $\sigma_j^i = -\left(\frac{1}{2\eta} - L_\theta\right)\|\theta_{j+1}^i - \theta_j^i\|^2 + \frac{1}{L_\theta}\|g_j^i - \nabla_\theta f(\mathbf{J}(\theta_j^i))\|^2$. Unrolling the recursive inequality showed in Lemma F.2 and taking the expectation of both sides, we obtain

$$\mathbb{E}\left[f(\mathbf{J}(\theta^*)) - f(\mathbf{J}(\theta_m^i))\right] \leq (1-\varepsilon)^m \mathbb{E}\left[f(\mathbf{J}(\theta^*)) - f(\mathbf{J}(\theta_0^i))\right] + \frac{(4L_\theta + \frac{2}{\eta})l_\theta^2}{M(1-\gamma)^2}\varepsilon$$
$$+ \sum_{j=0}^{m-1}(1-\varepsilon)^{m-j-1}\mathbb{E}\left[\sigma_j^i\right]$$

for all $\varepsilon \leq \bar{\varepsilon}$. Moreover, substituting $\varepsilon \to \frac{\varepsilon}{M^2 \ln(\varepsilon^{-1})}$, we arrive at

$$\mathbb{E}\left[f(\mathbf{J}(\theta^*)) - f(\mathbf{J}(\theta_m^i))\right] \leq \left(1 - \frac{\varepsilon}{M^2 \ln(\varepsilon^{-1})}\right)^m \mathbb{E}\left[f(\mathbf{J}(\theta^*)) - f(\mathbf{J}(\theta_0^i))\right]$$
$$+ \frac{(4L_\theta + \frac{2}{\eta})l_\theta^2}{M(1-\gamma)^2}\frac{\varepsilon}{\ln(\varepsilon^{-1})} + \sum_{j=0}^{m-1}(1 - \frac{\varepsilon}{M^2 \ln(\varepsilon^{-1})})^{m-j-1}\mathbb{E}\left[\sigma_j^i\right]$$

for all $\varepsilon \leq \bar{\varepsilon}$.

Next, note that by choosing $m = M^2\varepsilon^{-1}\ln(\varepsilon^{-1}), B = M^3\varepsilon^{-1}\ln(\varepsilon^{-1}), N = M^5\varepsilon^{-2}(\ln(\varepsilon^{-1}))^2$ and $\eta = \frac{1}{2L_\theta + 8C_3/(ML_\theta)}$ and if $\varepsilon \leq \frac{1}{2}$, we compute

$$\sum_{j=0}^{m-1}(1 - \frac{\varepsilon}{M^2\ln(\varepsilon^{-1})})^{m-j-1}\mathbb{E}\left[\sigma_j^i\right]$$

$$\leq \frac{1}{L_\theta}\sum_{j=0}^{m-1}\mathbb{E}\left[\|g_j^i - \nabla_\theta f(\mathbf{J}(\theta_j^i))\|^2\right]$$

$$- \left(1 - \frac{\varepsilon}{M^2\ln(\varepsilon^{-1})}\right)^{M^2\varepsilon^{-1}\ln(\varepsilon^{-1})}\left(\frac{1}{2\eta} - L_\theta\right)\sum_{j=0}^{m-1}\mathbb{E}\left[\|\theta_{j+1}^i - \theta_j^i\|^2\right]$$

$$\overset{(i)}{\leq} \frac{1}{L_\theta}\sum_{j=0}^{m-1}\mathbb{E}\left[\|g_j^i - \nabla_\theta f(\mathbf{J}(\theta_j^i))\|^2\right] - \left(\frac{1}{8\eta} - \frac{L_\theta}{4}\right)\sum_{j=0}^{m-1}\mathbb{E}\left[\|\theta_{j+1}^i - \theta_j^i\|^2\right]$$

$$\overset{(ii)}{\leq} \frac{mC_1}{L_\theta N} + \frac{mC_2\gamma^{2H}}{L_\theta} - \left(\frac{1}{8\eta} - \frac{L_\theta}{4} - \frac{mC_3}{L_\theta B}\right)\sum_{j=0}^{m-1}\mathbb{E}\left[\|\theta_{j+1}^i - \theta_j^i\|^2\right]$$

$$= \frac{C_1}{L_\theta M^3}\frac{\varepsilon}{\ln(\varepsilon^{-1})} + \frac{M^2 C_2 \ln(\varepsilon^{-1})}{L_\theta\varepsilon}\gamma^{2H}$$

where step (i) holds since $\left(1 - \frac{\varepsilon}{M^2\ln(\varepsilon^{-1})}\right)^{M^2\varepsilon^{-1}\ln(\varepsilon^{-1})} \geq \frac{1}{4}$ and step (ii) follows by Lemma 5.

Therefore, noting that $\left(1 - \frac{\varepsilon}{M^2\ln(\varepsilon^{-1})}\right)^{M^2\varepsilon^{-1}\ln(\varepsilon^{-1})} \leq \exp(-1)$ for $\varepsilon \in [0, \frac{1}{2}]$, we have

$$\mathbb{E}\left[f(\mathbf{J}(\theta^*)) - f(\mathbf{J}(\theta_m^i))\right] \leq \exp(-1)\mathbb{E}\left[f(\mathbf{J}(\theta^*)) - f(\mathbf{J}(\theta_0^i))\right] \tag{7}$$
$$+ \left(\frac{(4L_\theta + \frac{2}{\eta})l_\theta^2}{M(1-\gamma)^2} + \frac{C_1}{M^3 L_\theta}\right)\frac{\varepsilon}{\ln(\varepsilon^{-1})} + \frac{M^2 C_2 \ln(\varepsilon^{-1})}{L_\theta \varepsilon}\gamma^{2H}.$$

Unrolling the recursion given by inequality 7, we obtain

$$\mathbb{E}\left[f(\mathbf{J}(\theta^*)) - f(\mathbf{J}(\tilde{\theta}_T))\right] \leq \exp(-T)\mathbb{E}\left[f(\mathbf{J}(\theta^*)) - f(\mathbf{J}(\tilde{\theta}_0))\right]$$
$$+ \left(\frac{(4L_\theta + \frac{2}{\eta})l_\theta^2}{M(1-\gamma)^2} + \frac{C_1}{M^3 L_\theta}\right)\frac{T\varepsilon}{\ln(\varepsilon^{-1})} + \frac{TM^2 C_2 \ln(\varepsilon^{-1})}{L_\theta \varepsilon}\gamma^{2H}.$$

Finally, choosing $T = \ln(\varepsilon^{-1})$ and $H = \frac{\ln(M\varepsilon^{-1})}{1-\gamma}$, we conclude $\mathbb{E}\left[f(\mathbf{J}(\theta^*)) - f(\mathbf{J}(\tilde{\theta}_T))\right] \leq \mathcal{O}(\varepsilon)$, as desired. $\square$

## H  Experiments and Hardware Specifications

Our experiments were conducted on a high performance cluster. Specifically, we used 8 cores of a **96-core AMD EPYC 9654 processors** (2.4 GHz nominal, 3.7 GHz peak), with 10 GB RAM per core from a **384 GB of DDR5 memory** clocked at 4800 MHz.

The experiments were executed utilizing 8 CPU cores, with the jobs distributed across the cores in parallel. Each training run was performed over a total of 2400 epochs, and the process involved 8 independent runs to ensure robustness of the results. The training of the multi-objective reinforcement learning algorithms, specifically MOPG and MOTSIVRPG, spanned approximately 2.5 days per run for the server queues environment. Intermediate results were checkpointed every 200 epochs to facilitate progress monitoring and potential recovery.

## I  Additional Experiments

The following plots are part of the experiments conducted in Section 5 to validate the theoretical guarantees on sample complexity. For space reasons, and because none of them adds anything to the conclusions we were already able to extrapolate, it was not possible nor necessary to include any of these in the main corpse of the paper.

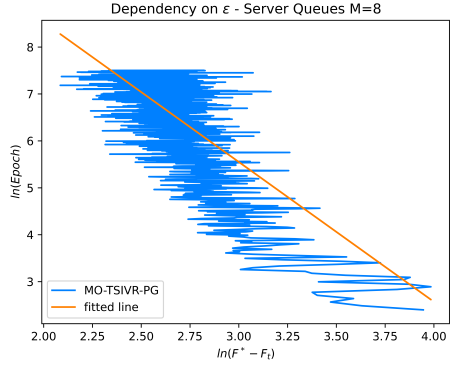

Figure 5: Log plot of the optimality gap against the number of epochs on the Server Queues environment with $M = 8$

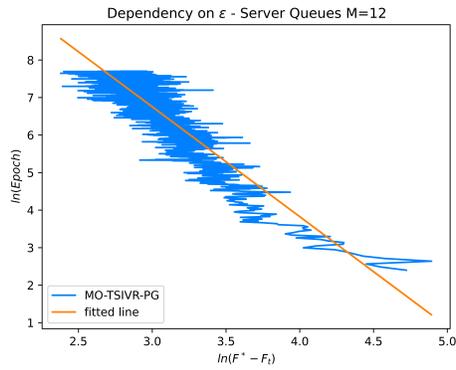

Figure 6: Log plot of the optimality gap against the number of epochs on the Server Queues environment with $M = 12$

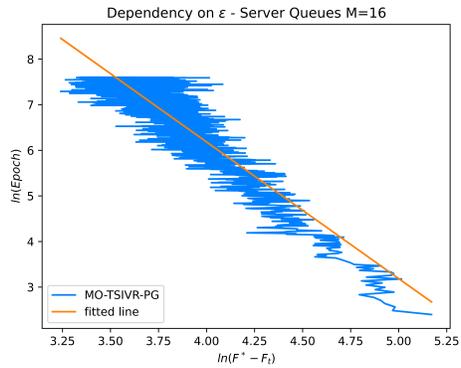

Figure 7: Log plot of the optimality gap against the number of epochs on the Server Queues environment with $M = 16$

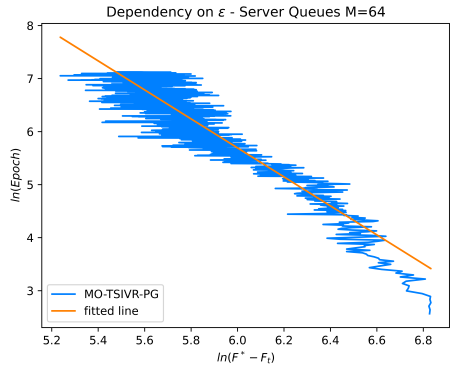

Figure 8: Log plot of the optimality gap against the number of epochs on the Server Queues environment with $M = 64$

## J  Reproducibility Checklist

This paper:

- Includes a conceptual outline and/or pseudocode description of AI methods introduced (yes)
- Clearly delineates statements that are opinions, hypothesis, and speculation from objective facts and results (yes)
- Provides well marked pedagogical references for less-familiare readers to gain background necessary to replicate the paper (no)

Does this paper make theoretical contributions? (yes)
If yes, please complete the list below.

- All assumptions and restrictions are stated clearly and formally. (yes)
- All novel claims are stated formally (e.g., in theorem statements). (yes)
- Proofs of all novel claims are included. (yes)
- Proof sketches or intuitions are given for complex and/or novel results. (no, we try to provide the reader with an intuition to our contributions, but proofs of our main results are analyzed formally in the appendix)
- Appropriate citations to theoretical tools used are given. (NA)
- All theoretical claims are demonstrated empirically to hold. (partial)
- All experimental code used to eliminate or disprove claims is included. (NA)

Does this paper rely on one or more datasets? (no)
If yes, please complete the list below.

- A motivation is given for why the experiments are conducted on the selected datasets (yes/partial/no/NA)
- All novel datasets introduced in this paper are included in a data appendix. (yes/partial/no/NA)
- All novel datasets introduced in this paper will be made publicly available upon publication of the paper with a license that allows free usage for research purposes. (yes/partial/no/NA)
- All datasets drawn from the existing literature (potentially including authors' own previously published work) are accompanied by appropriate citations. (yes/no/NA)
- All datasets drawn from the existing literature (potentially including authors' own previously published work) are publicly available. (yes/partial/no/NA)
- All datasets that are not publicly available are described in detail, with explanation why publicly available alternatives are not scientifically satisficing. (yes/partial/no/NA)

Does this paper include computational experiments? (yes) If yes, please complete the list below.

- Any code required for pre-processing data is included in the appendix. (yes).
- All source code required for conducting and analyzing the experiments is included in a code appendix. (yes)
- All source code required for conducting and analyzing the experiments will be made publicly available upon publication of the paper with a license that allows free usage for research purposes. (yes)
- All source code implementing new methods have comments detailing the implementation, with references to the paper where each step comes from (no)
- If an algorithm depends on randomness, then the method used for setting seeds is described in a way sufficient to allow replication of results. (yes)
- This paper specifies the computing infrastructure used for running experiments (hardware and software), including GPU/CPU models; amount of memory; operating system; names and versions of relevant software libraries and frameworks. (yes)
- This paper formally describes evaluation metrics used and explains the motivation for choosing these metrics. (yes)
- This paper states the number of algorithm runs used to compute each reported result. (yes)
- Analysis of experiments goes beyond single-dimensional summaries of performance (e.g., average; median) to include measures of variation, confidence, or other distributional information. (yes)
- The significance of any improvement or decrease in performance is judged using appropriate statistical tests (e.g., Wilcoxon signed-rank). (no)
- This paper lists all final (hyper-)parameters used for each model/algorithm in the paper's experiments. (yes)
- This paper states the number and range of values tried per (hyper-) parameter during development of the paper, along with the criterion used for selecting the final parameter setting. (no)